\documentclass[11pt,letterpaper,onecolumn]{article}

\usepackage[T1]{fontenc}
\usepackage[utf8]{inputenc}
\usepackage{times}
\usepackage[letterpaper,textwidth=6.5in,textheight=9in,centering,
  headheight=14pt,headsep=18pt,footskip=28pt]{geometry}
\usepackage{natbib}
\setcitestyle{authoryear,round,citesep={;},aysep={,},yysep={;}}
\usepackage{hyperref}
\usepackage{url}
\usepackage{pifont}
\usepackage{xcolor}
\usepackage{colortbl}
\usepackage{makecell}
\usepackage{booktabs}
\definecolor{good}{RGB}{34,139,34}
\definecolor{bad}{RGB}{178,34,34}
\definecolor{mid}{RGB}{218,150,40}
\definecolor{seerbg}{RGB}{214,234,248}
\definecolor{headbg}{RGB}{240,240,240}
\definecolor{EviBlue}{HTML}{6C8EBF}
\definecolor{EviCoral}{HTML}{B85450}
\definecolor{EviBlueTint}{HTML}{EDF3F9}
\definecolor{EviCoralTint}{HTML}{F9EFEE}
\definecolor{EviLavenderTint}{HTML}{F1EEFA}
\definecolor{EviMintTint}{HTML}{EAF6F1}
\definecolor{EviBlueInk}{HTML}{365F8F}
\definecolor{EviCoralInk}{HTML}{A3423D}

\definecolor{improvement}{RGB}{34,139,34}
\definecolor{decline}{RGB}{220,20,60}

\usepackage{microtype}
\usepackage{multirow}
\usepackage{booktabs}
\usepackage{caption}
\usepackage{pgfplots}
\pgfplotsset{compat=1.18}
\usepackage{graphicx}
\usepackage{subcaption}
\usepackage{enumitem}
\usepackage{float}
\usepackage{wrapfig}
\usepackage{amsmath, amssymb}

\usepackage{tabularx}
\usepackage[most]{tcolorbox}
\usepackage{array}
\newcolumntype{L}[1]{>{\raggedright\arraybackslash}p{#1}}
\usepackage{makecell}
\usepackage{tikz}
\usepackage{algorithm2e}

\usepackage{colortbl}
\definecolor{cellgray}{gray}{0.85}
\definecolor{textgray}{gray}{0.85}
\definecolor{R_red}{HTML}{F8CECC}

\definecolor{red4}{RGB}{133, 193, 233}
\definecolor{red3}{RGB}{174, 214, 241}
\definecolor{red2}{RGB}{214, 234, 248}
\definecolor{red1}{RGB}{235, 245, 251}
\definecolor{green1}{RGB}{247, 220, 111}
\definecolor{green2}{RGB}{249, 231, 159}
\definecolor{green3}{RGB}{252, 243, 207}
\definecolor{green4}{RGB}{254, 249, 231}

\usepackage{amsthm}
\theoremstyle{plain}

\theoremstyle{definition}

\usepackage{xcolor}
\usepackage{listings} 
\newtcolorbox{modeloutput}[2][]{
  colback=gray!3,
  colframe=gray!80,
  title=\textbf{#2},
  fonttitle=\bfseries,
  sharp corners,
  boxrule=0.8pt,
  left=6pt,right=6pt,top=6pt,bottom=6pt,
  fontupper=\ttfamily\small,
  breakable,
  #1
}

\usepackage{adjustbox}

\definecolor{slotENTfg}{HTML}{D95B1A}
\definecolor{slotENTbg}{HTML}{FDF0E8}
\definecolor{slotATTfg}{HTML}{F0A500}
\definecolor{slotATTbg}{HTML}{FEF8E7}
\definecolor{slotRELfg}{HTML}{2E9E5B}
\definecolor{slotRELbg}{HTML}{EBF8F0}
\definecolor{slotDETfg}{HTML}{2369A8}
\definecolor{slotDETbg}{HTML}{EBF3FB}
\definecolor{slotSUMfg}{HTML}{B02020}
\definecolor{slotSUMbg}{HTML}{FDECEA}

\usepackage{fancyhdr}
\usepackage{needspace}
\usepackage{etoolbox}

\hypersetup{
  hidelinks,
  pdftitle={Learning Multimodal Embeddings with Evidence-Aligned Readout},
  pdfauthor={Zirong Chen, Fuda Ye, Enjun Du, Junfu Pu, Xinlei Wang, Xinyu Zuo, Lisheng Duan, Haijin Liang, Jin Ma, Jiachuan Wang, Yongqi Zhang},
  pdfsubject={Multimodal retrieval with evidence-aligned readout}
}

\fancypagestyle{plain}{\fancyhf{}
  \fancyhead[L]{\small Yuanbao Technical Report}
  \fancyhead[R]{\small Tencent}
  \fancyfoot[C]{\thepage}
  
}

\pretocmd{\subsection}{\Needspace{4\baselineskip}}{}{}
\renewenvironment{abstract}{\begin{center}\bfseries Abstract\end{center}
  \begin{list}{}{\setlength{\leftmargin}{0.2in}\setlength{\rightmargin}{0.2in}}
  \item\relax\small
}{\end{list}\vspace{0.5em}}

\begin{document}
\noindent
\begin{minipage}[c]{0.45\linewidth}
  \includegraphics[width=0.40\linewidth]{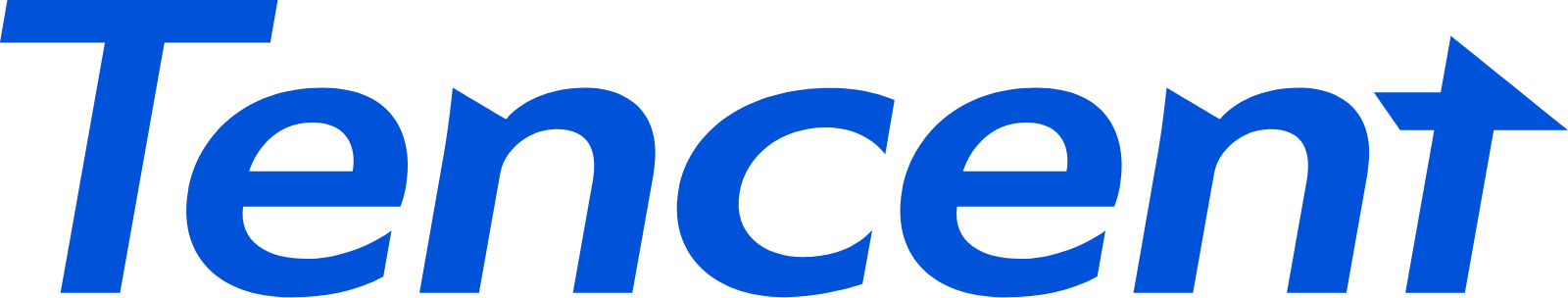}
\end{minipage}\hfill
\begin{minipage}[c]{0.45\linewidth}
  \raggedleft\includegraphics[height=1.25cm]{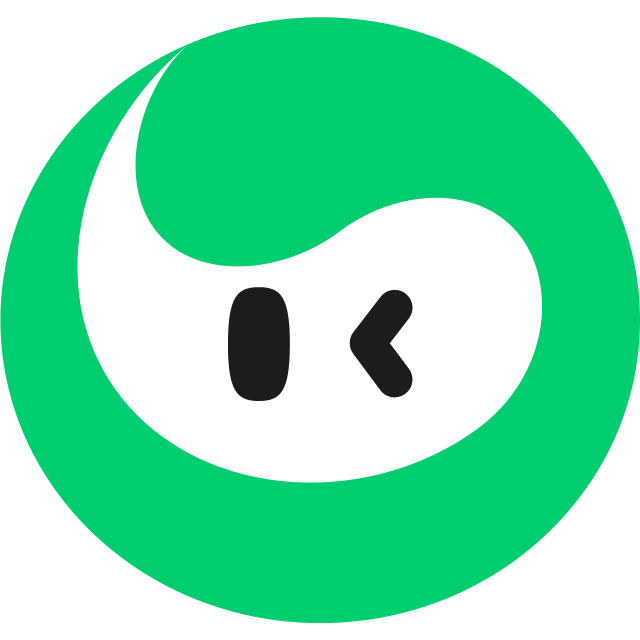}
\end{minipage}

\vspace{0.35cm}
\begin{center}
  {\fontsize{22}{26}\selectfont\bfseries
  Learning Multimodal Embeddings\\[0.12em]
  with Evidence-Aligned Readout\par}

  \vspace{0.4cm}
  {\normalsize
  Zirong Chen\textsuperscript{1,2,3,\textdagger,\textdaggerdbl} \quad
  Fuda Ye\textsuperscript{1,\textdaggerdbl} \quad
  Enjun Du\textsuperscript{1,2,5,\textdagger} \quad
  Junfu Pu\textsuperscript{4}\\[0.35em]
  Xinlei Wang\textsuperscript{2} \quad
  Xinyu Zuo\textsuperscript{2} \quad
  Lisheng Duan\textsuperscript{2} \quad
  Haijin Liang\textsuperscript{2}\\[0.35em]
  Jin Ma\textsuperscript{2} \quad
  Jiachuan Wang\textsuperscript{6} \quad
  Yongqi Zhang\textsuperscript{1,*}\par}

  \vspace{0.3cm}
  {\small
  \textsuperscript{1}The Hong Kong University of Science and Technology (Guangzhou)\\[0.15em]
  \textsuperscript{2}Tencent Yuanbao \qquad
  \textsuperscript{3}Tsinghua University\\[0.15em]
  \textsuperscript{4}ARC Lab, Tencent \qquad
  \textsuperscript{5}The University of Hong Kong\\[0.15em]
  \textsuperscript{6}University of Tsukuba\\[0.25em]
  \texttt{\href{mailto:imzrchen@gmail.com}{imzrchen@gmail.com},
  \href{mailto:yongqizhang@hkust-gz.edu.cn}{yongqizhang@hkust-gz.edu.cn}}\par}
\end{center}
\vspace{0.15cm}
\begingroup
  \renewcommand{\thefootnote}{\fnsymbol{footnote}}
  \footnotetext[2]{Work done during an internship at Tencent.}
  \footnotetext[3]{Zirong Chen and Fuda Ye contributed equally.}
  \footnotetext[1]{Corresponding author.}
\endgroup
\setcounter{footnote}{0}

\begin{abstract}

Multimodal large language models can expose task-relevant evidence through generation, but producing useful evidence does not by itself determine how it enters a retrieval embedding.
We study whether the semantic organization of that evidence can also specify where representations are read.
To address this question, we introduce \textbf{EviAlign}, which couples \emph{Semantic Evidence Generation} with \emph{Boundary Readout} in a shared multimodal large language model.
It organizes evidence into five semantic units, reads the contextualized state at each unit boundary, and aggregates these states into a single normalized embedding.
Generation and contrastive retrieval objectives jointly train this shared structure.
With the same trailing readout, semantic evidence and free-form CoT yield nearly identical retrieval performance, suggesting that evidence organization alone does not explain the full gain.
A controlled $2\times3$ study compares consistent and permuted evidence organization across three readout strategies, using training targets with matched evidence spans.
With five readout states and the same mean pooling, the advantage of consistent semantic organization grows from 0.65 points at length-based training positions to 2.39 at evidence boundaries, yielding a 1.74-point co-design interaction.
Across 12 MMEB retrieval tasks, EviAlign achieves 76.9 average Recall@1 with 500K training pairs while retaining single-vector indexing and scoring.

\end{abstract}

\section{Introduction}
\label{sec:introduction}
Multimodal embedding models map text, images, and interleaved inputs into a shared space for retrieval~\citep{Jiang2024VLM2VecTV,Zhang2024GMEIU}. Across tasks, a relevant match may depend on different aspects of the input, including entities, attributes, relations, and task-specific details. Multimodal large language models (MLLMs) can expose these cues through generation. We refer to such task-relevant cues as \emph{retrieval evidence} and study how they can be incorporated into a single retrieval embedding. For example, a composed-image query can pair a photograph of one bottle with an instruction to retrieve three. The relevant evidence must express the requested count and arrangement while preserving identifying visual cues from the photograph. The central challenge is that generating useful evidence does not determine how it contributes to the embedding.

Recent work introduces explicit reasoning, generated context, retrieval-oriented rewrites, or latent computation before embedding~\citep{Lan2025UMER1ER,Cui2025ThinkTE,Wu2026RIME,Wu2026LaME}, while flexible readouts use learned aggregation or compact multi-vector representations~\citep{Cui2025ThinkTE,Xiao2025MetaEmbedSM,Faysse2024ColPaliED}. Some of these systems jointly optimize generation and embedding. We study a more specific design question: \emph{Can the semantic units produced during generation also specify where retrieval representations are read?}

\begin{figure}[t]
    \centering
    \includegraphics[width=\linewidth]{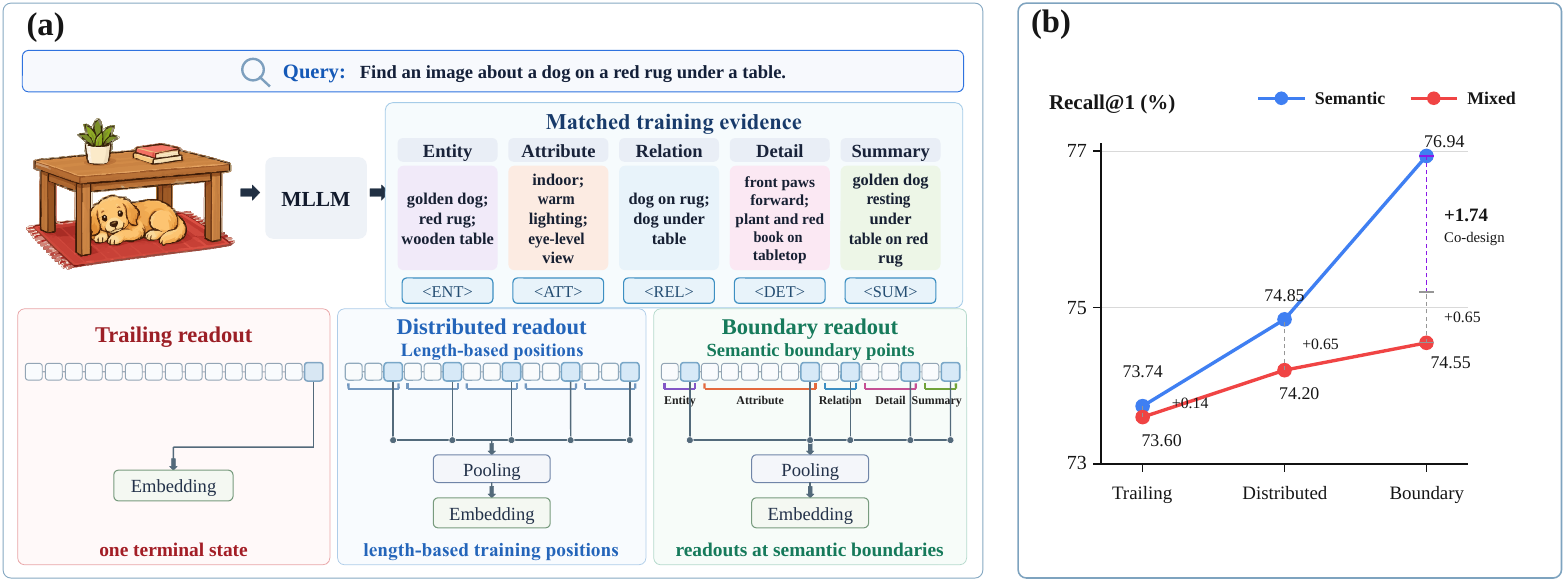}
    \caption{\textbf{Evidence--readout co-design in EviAlign.}
    \textbf{(a)} Training targets share five evidence spans across readout conditions. Distributed uses length-based training positions; at inference, readouts follow the emitted tokens.
    \textbf{(b)} Mixed permutes labeled evidence spans while fixing boundary-token identities and order; Semantic preserves role-to-boundary correspondence. Boundary readout enlarges the semantic--mixed gap from 0.65 to 2.39 points, yielding a 1.74-point co-design interaction.}
    \label{fig:intro_challenge}
\end{figure}

Figure~\ref{fig:intro_challenge}(a) illustrates three ways to construct an embedding from generated evidence: reading a trailing state, pooling distributed states whose locations are independent of evidence boundaries, and pooling states at evidence-unit boundaries. In a controlled study whose Semantic and Mixed training targets contain the same evidence spans, semantic organization yields its largest advantage when readouts follow these boundaries (Figure~\ref{fig:intro_challenge}(b)). This motivates \emph{evidence-aligned readout}: using the semantic organization of generated evidence to determine where contextualized states are read for embedding construction.

In this paper, we introduce \textbf{EviAlign}, which couples \emph{Semantic Evidence Generation} with \emph{Boundary Readout} in a shared MLLM. It organizes retrieval evidence into five semantic units---Entity, Attribute, Relation, Detail, and Summary---and aggregates the contextualized states at their boundaries into a single normalized embedding. Each boundary state is read after its evidence unit has been completed, with access to the input and preceding evidence. Generation supervision and contrastive retrieval jointly optimize the model.

A controlled $2\times3$ study yields a \textbf{1.74-point co-design interaction}; readout analyses further characterize the shared and complementary information retained by the boundary states. On 12 MMEB retrieval tasks, EviAlign reaches \textbf{76.9} average Recall@1 with \textbf{500K training pairs}. These results support using evidence structure as an interface for representation construction while retaining conventional single-vector indexing and scoring.

Overall, our contributions are as follows:
\begin{itemize}[nosep]
    \item We formulate \emph{evidence-aligned readout}: the semantic organization of generated evidence explicitly determines the roles and locations of representation readouts.

    \item We introduce \textbf{EviAlign}, which implements this principle through Semantic Evidence Generation and Boundary Readout while retaining a single-vector indexing and scoring interface.

    \item Through matched controls, we show that the benefit of semantic evidence depends on how its states are read: neither semantic generation with a trailing readout nor additional readout tokens alone reproduces the full gain.
\end{itemize}

\section{Related Work}
\paragraph{Multimodal Embedding Learning.}

Contrastive vision--language models such as CLIP~\citep{Radford2021LearningTV}, BLIP~\citep{Li2022BLIPBL}, and SigLIP~\citep{Zhai2023SigmoidLF} learn shared representation spaces for cross-modal matching. Building on MLLMs, unified multimodal embedders support task-conditioned retrieval across text, images, and interleaved multimodal inputs~\citep{Jiang2024VLM2VecTV,Lin2024MMEmbedUM,Zhang2024GMEIU}. UniME-V2 and Qwen3-VL-Embedding further scale direct multimodal embedding through stronger supervision and multi-stage, multi-source training; Qwen3-VL-Embedding reports 80.2 on MMEB-V2 Image RET~\citep{Gu2025UniMEV2MF,li2026qwen3vlembedding}. These systems construct a retrieval representation directly from contextualized model states, without using generated semantic units to specify its readout locations.

Complementary to embedding-model development, SnapBench evaluates snap-and-ask retrieval under paired visual and textual corruptions~\citep{chen2026snapbench}. Its analysis highlights the importance of distinguishing the contributions of the query's visual content and textual request when constructing a retrieval representation.

\paragraph{Generation-Enhanced Embedding Learning.}

Explicit generation also supports multimodal reasoning: SOPHIA develops vision--language slow-thinking through semi-off-policy reinforcement learning~\citep{NEURIPS2025_b5dc49f4}. Generation-enhanced retrieval introduces a further requirement: converting the generated context into a representation suitable for matching.

Generation-enhanced embedders expose retrieval-relevant information before constructing the final representation. UME-R1 and Think-Then-Embed generate reasoning traces or task-conditioned context, while Reasoning Guided Embeddings and RIME produce retrieval-oriented descriptions or rewrites~\citep{Lan2025UMER1ER,Cui2025ThinkTE,Liu2025ReasoningGE,Wu2026RIME}. PLUME and LaME instead perform embedding-oriented computation in latent states rather than readable textual evidence~\citep{He2026PLUME,Wu2026LaME}. Several of these methods jointly optimize generation and embedding. EviAlign focuses on a different structural question: whether generated semantic units can explicitly determine the roles and locations of the states used to construct the retrieval embedding.

\paragraph{Structured Evidence in Multimodal Decisions.}

Structured evidence serves several distinct computational roles. EviRank represents multimodal relevance as typed constraints and verifies candidate images against them for re-ranking~\citep{du2026evirank}. LEDGERMIND maintains a provenance-constrained evidence ledger for multi-step multimodal reasoning~\citep{du2026ledgermind}, while Omni-Streaming Thinking separates observed evidence from forecasts and verifies claims as audio-visual streams unfold~\citep{du2026omni}. These systems make evidence explicit for verification and decision-making. EviAlign connects evidence organization to representation construction: the generated units define the boundary states jointly trained and aggregated for retrieval.

\paragraph{Representation Readout for Retrieval.}

In whole-slide imaging, DRE-SLCL aggregates tile features through dynamic residual encoding and trains slide representations with a slide-level contrastive objective~\citep{jin2025dynamic}. This provides a domain-specific example of coupling local-feature aggregation with global representation learning.

Readout locations may be trailing, learned implicitly, or associated with predefined spans. Most single-vector MLLM embedders use a trailing hidden state~\citep{Jiang2024VLM2VecTV}; other designs apply learned attention, latent-context pooling, or query-based aggregation to contextualized features~\citep{Cui2025ThinkTE}. Late chunking constructs representations from contextualized token states over predefined text spans~\citep{Gunther2024LateChunking}. Boundary Readout uses boundaries defined by generated, task-relevant semantic evidence. Multi-vector systems such as ColBERT, ColPali, and MetaEmbed retain several representations for late interaction~\citep{Khattab2020ColBERTEA,Faysse2024ColPaliED,Xiao2025MetaEmbedSM}; EviAlign aggregates its internal readouts into one vector for standard indexing and similarity scoring.

\section{EviAlign}
\label{sec:method}

\begin{figure}[t]
    \centering
    \includegraphics[width=0.99\linewidth]{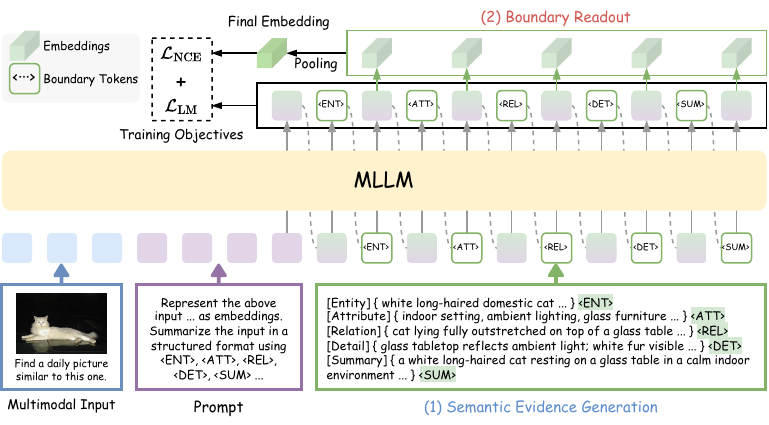}
    \caption{\textbf{Overview of EviAlign's evidence--readout co-design.}
    Semantic Evidence Generation defines evidence units and their boundaries; Boundary Readout extracts states at those boundaries and pools them into one normalized embedding.}
    \label{fig:overview}
\end{figure}

We introduce \textbf{EviAlign}, a generation-assisted multimodal embedding framework that co-designs evidence generation and representation readout through a shared boundary interface. As shown in Figure~\ref{fig:overview}, Semantic Evidence Generation produces task-relevant units with explicit boundaries, and Boundary Readout extracts and aggregates the contextualized states at those boundaries into a single retrieval embedding. A shared multimodal large language model (MLLM) performs both operations.

\subsection{Problem Formulation and Design Principle}
\label{sec:problem_formulation}

Let $\mathcal{D}=\{(q_i,c_i^{+})\}_{i=1}^{|\mathcal{D}|}$ denote a collection of query--candidate training pairs, where $c_i^{+}$ is relevant to query $q_i$. A query or candidate may contain text, an image, an interleaved image--text input, and its associated task instruction. We learn a shared encoder
\begin{equation}
    \mathbf{h}_{\theta}(x)\in\mathbb{R}^{d},
    \qquad
    \|\mathbf{h}_{\theta}(x)\|_{2}=1,
    \label{eq:embedding_function}
\end{equation}
and rank candidates using
\begin{equation}
    s_{\theta}(q,c)
    =
    \mathbf{h}_{\theta}(q)^{\top}\mathbf{h}_{\theta}(c).
    \label{eq:retrieval_similarity}
\end{equation}
Queries and candidates are independently encoded by the same model. In particular, candidate representations are not conditioned on the query against which they are subsequently compared, allowing them to be precomputed and indexed.

A generation-assisted embedding model can be expressed as
\begin{equation}
    \mathbf{z}=G_{\theta}\!\left(\texttt{P}(x)\right),
    \qquad
    \mathbf{h}_{\theta}(x)=R_{\theta}\!\left(\texttt{P}(x),\mathbf{z}\right),
    \label{eq:generation_assisted_embedding}
\end{equation}
where $\texttt{P}(x)$ combines the multimodal input with its task instruction, $G_{\theta}$ denotes autoregressive generation, and $R_{\theta}$ denotes representation readout. A single MLLM performs both operations. Equations~\ref{eq:evidence_slot_readout}--\ref{eq:evidence_aggregation} below instantiate $R_{\theta}$ for EviAlign.

Equation~\ref{eq:generation_assisted_embedding} exposes two coupled design choices: how retrieval evidence is expressed during generation and how that evidence contributes to the resulting embedding. These choices interact: richer generated content need not preserve its organization in a trailing readout, and additional readout tokens need not align with evidence units.

We therefore formulate \emph{evidence-aligned readout} as a co-design principle with three requirements. First, generation should select task-relevant evidence that supports the intended match or distinguishes it from misleadingly similar candidates. Second, the organization of this evidence should explicitly determine the semantic roles and locations of representation readouts. Third, the resulting readouts should be jointly integrated into an embedding optimized for retrieval.

EviAlign realizes these requirements through two components that share the same evidence structure. Semantic Evidence Generation defines the evidence units and their boundaries, while Boundary Readout uses these boundaries as representation readout locations. Their aggregation yields one normalized vector per input.

\subsection{Semantic Evidence Generation}
\label{sec:method_evidence_generation}

Semantic Evidence Generation selects task-relevant information and organizes it into units that directly guide representation readout. The generated content focuses on evidence that affects relevance under the given instruction. For composed retrieval, for example, the prompt contains both the reference input and the requested modification, and the generated evidence describes the intended target after applying that modification rather than merely captioning the reference image.

Given an input $x$, the MLLM generates a token sequence partitioned into evidence spans and their following boundary tokens:
\begin{equation}
    \mathbf{z}
    =
    (\mathbf{z}_1,s_1,\mathbf{z}_2,s_2,\ldots,\mathbf{z}_K,s_K),
    \label{eq:structured_evidence}
\end{equation}
where $\mathbf{z}$ denotes the complete generated token sequence, $\mathbf{z}_k$ is the contiguous span of the $k$-th task-relevant evidence unit, and $s_k$ is a boundary token placed immediately after that span. Each evidence-unit type specifies a semantic role; the following token $s_k$ marks the end of the unit, and its final-layer state serves as the corresponding boundary readout.

In our implementation, $K=5$: Entity ends with \texttt{<ENT>}, Attribute with \texttt{<ATT>}, Relation with \texttt{<REL>}, Detail with \texttt{<DET>}, and Summary with \texttt{<SUM>}. Entity records objects or concepts together with identifying properties; Attribute captures scene-level characteristics; Relation describes actions, interactions, and spatial relations; Detail retains locally discriminative cues such as visible text and fine-grained patterns; and Summary provides a retrieval-focused global description.

The boundary tokens are new vocabulary items initialized from the \texttt{[EOS]} embedding. Their fixed role-to-boundary correspondence lets the same structure organize evidence and provide readout states. Each query or candidate is encoded with its own task instruction, following the independent encoding interface in Section~\ref{sec:problem_formulation}. Section~\ref{sec:representation_analysis} evaluates joint evidence-schema and readout configurations; Appendix~\ref{app:prompt} provides the complete generation prompt.

\subsection{Boundary Readout}
\label{sec:method_evidence_extract}

Boundary Readout makes this evidence structure part of embedding construction by forming a contextualized representation at each boundary defined in Equation~\ref{eq:structured_evidence}.

Let $p_k$ denote the position of $s_k$ in the generated sequence, with the input-prefix offset implicit. We extract
\begin{equation}
    \mathbf{e}_k
    =
    H_{\theta}\!\left(\texttt{P}(x),\mathbf{z}_{\leq p_k}\right)[p_k],
    \qquad k=1,\ldots,K,
    \label{eq:evidence_slot_readout}
\end{equation}
where $H_{\theta}(\cdot)[p_k]$ denotes the final-layer hidden state at position $p_k$. Under causal attention, $s_k$ is the first designated readout position after the complete unit $\mathbf{z}_k$, so $\mathbf{e}_k$ summarizes the input-conditioned evidence prefix available at that position rather than only the immediately preceding unit. The semantic roles specify where representations are read, while retrieval training determines how these contextualized states jointly contribute to the final embedding.

We aggregate the evidence-aligned readouts and normalize the result:
\begin{equation}
    \bar{\mathbf{h}}_{\theta}(x)
    =
    \frac{1}{K}\sum_{k=1}^{K}\mathbf{e}_k,
    \qquad
    \mathbf{h}_{\theta}(x)
    =
    \frac{\bar{\mathbf{h}}_{\theta}(x)}
         {\|\bar{\mathbf{h}}_{\theta}(x)\|_2}.
    \label{eq:evidence_aggregation}
\end{equation}
Mean pooling assigns equal coefficients to the five contextualized boundary states, and L2 normalization produces a single cosine-scored vector for indexing and retrieval.

Here \emph{alignment} denotes the within-input correspondence between evidence units and representation readout locations. Retrieval operates on the aggregated global embeddings in Equation~\ref{eq:retrieval_similarity}.

\subsection{Training and Inference}
\label{sec:optimization}

EviAlign jointly learns evidence generation and retrieval representation construction. During supervised training, each input $x$ is associated with a target semantic-evidence sequence $\mathbf{y}=(y_1,\ldots,y_T)$, which contains both the evidence content and its boundary tokens. Under teacher forcing, the representation associated with the $k$-th evidence boundary is
\begin{equation}
    \mathbf{e}_k^{\mathrm{train}}
    =
    H_{\theta}\!\left(\texttt{P}(x),\mathbf{y}_{\leq p_k(\mathbf{y})}\right)[p_k(\mathbf{y})].
    \label{eq:teacher_forced_readout}
\end{equation}
Here $p_k(\mathbf{y})$ is the boundary-token position in the teacher-forced target, rather than in a generated sequence. These readouts are aggregated by Equation~\ref{eq:evidence_aggregation}.

Given a batch $\mathcal{B}=\{(q_i,c_i^{+})\}_{i=1}^{|\mathcal{B}|}$, we optimize a symmetric in-batch contrastive objective. Let $s_{ij}=\mathbf{h}_{\theta}(q_i)^{\top}\mathbf{h}_{\theta}(c_j^{+})/\tau$. The loss averages query-to-candidate and candidate-to-query retrieval:

\begin{equation}
    \mathcal{L}_{\text{NCE}}
    = -\frac{1}{2|\mathcal{B}|}\sum_{i=1}^{|\mathcal{B}|}
    \left[
        \log\frac{\exp(s_{ii})}{\sum_j\exp(s_{ij})}
        +
        \log\frac{\exp(s_{ii})}{\sum_j\exp(s_{ji})}
    \right],
    \label{eq:contrastive_loss}
\end{equation}
where $\tau$ is a temperature parameter. This objective trains the aggregated representation to preserve evidence that distinguishes the relevant candidate from in-batch negatives.

We sum the query- and candidate-side language-modeling losses, each averaged over non-masked supervised target tokens:
\begin{equation}
    \mathcal{L}_{\text{LM}} = \mathcal{L}_{\text{LM}}^q + \mathcal{L}_{\text{LM}}^c.
    \label{eq:lm_loss}
\end{equation}
The training objective is
\begin{equation}
    \mathcal{L} = \mathcal{L}_{\text{NCE}} + \mathcal{L}_{\text{LM}}.
    \label{eq:total_training_loss}
\end{equation}

The language-modeling objective teaches the model to express task-relevant evidence under the prescribed organization, whereas the contrastive objective trains the boundary readouts to form a discriminative retrieval embedding. Together they optimize the two sides of the evidence structure shared by generation and readout.

At inference time, no target evidence sequence is provided. The MLLM first autoregressively generates $\hat{\mathbf{z}}=G_{\theta}(\texttt{P}(x))$, after which the model reads the hidden states at the generated evidence boundaries and aggregates them into $\mathbf{h}_{\theta}(x)$. If a required boundary token is absent, its readout falls back to the final-layer hidden state of the last generated token. Queries and candidates follow the same generate-and-read procedure but are encoded independently. Candidate representations are computed offline and stored in the retrieval index, while query-side generation remains part of online encoding. Retrieval then uses Equation~\ref{eq:retrieval_similarity}. Algorithm~\ref{algo:evialign} in Appendix~\ref{app:algorithm} summarizes the complete procedure.

\section{Experiments}
\subsection{Experimental Setup}
\label{sec:exp_setup}

\paragraph{Datasets and Metrics.}
We focus training and evaluation on MMEB's 12 retrieval tasks, which directly match our study of representation construction for candidate ranking~\citep{Jiang2024VLM2VecTV}. Following the standard protocol, each dataset contains 1000 test queries, and each query is evaluated against a candidate pool of 1000 items (one positive and 999 negatives). The benchmark defines eight in-domain and four out-of-domain retrieval datasets, listed in Appendix~\ref{app:benchmarks}. We report Recall@1 and its average across all 12 datasets. With one positive per query, Recall@1 equals Precision@1 under this protocol.

\begin{table}[t]
    \caption{Comparison on the 12 MMEB retrieval tasks (Recall@1, \%). In-Domain and Out-of-Domain average eight and four tasks, respectively. Split averages are computed from the corresponding per-task results.$^{\dagger}$UniME-V2 and LaME report the 12-task retrieval mean without the 8/4 split. }
    \label{tab:main_results}
    \centering
    \scriptsize
    \setlength{\tabcolsep}{2.6pt}
    \renewcommand{\arraystretch}{1.06}
    \begin{NoHyper}
    \begin{tabular*}{\linewidth}{@{\extracolsep{\fill}}l l c c c c}
        \toprule
        \textbf{Method} & \textbf{Backbone} & \textbf{Data} & \textbf{In-Domain} & \textbf{Out-of-Domain} & \textbf{Overall} \\
        \midrule
        \rowcolor{EviBlueTint}
        \multicolumn{6}{l}{\textit{Direct embedding}} \\
        GME~\citep{Zhang2024GMEIU} & Qwen2-VL-7B & $\sim$8M & 70.9 & \textbf{71.8} & 71.2 \\
        LamRA-Ret~\citep{Liu2024LamRALM} & Qwen2-VL-7B & $\sim$1.4M & 70.0 & \underline{69.9} & 70.0 \\
        VLM2Vec~\citep{Jiang2024VLM2VecTV} & Qwen2-VL-7B & $\sim$662K & 75.2 & 57.9 & 69.4 \\
        VLM2Vec-V2~\citep{Meng2025VLM2VecV2AM} & Qwen2-VL-2B & $\sim$1.7M & 74.8 & 58.7 & 69.5 \\
        UniME-V2~\citep{Gu2025UniMEV2MF} & Qwen2-VL-7B & $\sim$662K & -- & -- & 73.1$^{\dagger}$ \\
        \midrule
        \rowcolor{EviCoralTint}
        \multicolumn{6}{l}{\textit{Explicit generation or reasoning}} \\
        UME-R1~\citep{Lan2025UMER1ER} & Qwen2-VL-7B & $\sim$1.5M & 75.4 & 64.8 & 71.9 \\
        RIME~\citep{Wu2026RIME} & Qwen2-VL-7B & $\sim$1.5M & 77.0 & 65.6 & 73.2 \\
        Think-Then-Embed$_t$~\citep{Cui2025ThinkTE} & Qwen2-VL-7B & $\sim$662K & -- & -- & \underline{75.9} \\
        \midrule
        \rowcolor{EviLavenderTint}
        \multicolumn{6}{l}{\textit{Latent reasoning}} \\
        PLUME~\citep{He2026PLUME} & Qwen2-VL-2B & $\sim$1.5M & 71.5 & 59.8 & 67.6 \\
        LaME~\citep{Wu2026LaME} & Qwen2-VL-7B & $\sim$1.55M & -- & -- & 73.1$^{\dagger}$ \\
        \midrule
        \rowcolor{EviMintTint}
        \textbf{EviAlign} & Qwen2-VL-7B & \textbf{500K} & \underline{80.7} & 65.1 & 75.5 \\
        \rowcolor{EviMintTint}
        \textbf{EviAlign} & Qwen3-VL-8B & \textbf{500K} & \textbf{81.6} & 67.7 & \textbf{76.9} \\
        \bottomrule
    \end{tabular*}
    \end{NoHyper}
    \vspace{2pt}
\end{table}

\paragraph{Implementation Details.}
Our main model and all controlled analyses reported in the main text use 500K query--candidate training pairs from the MMEB-V1 retrieval training split. Semantic-evidence targets for EviAlign are generated by GLM-4.1V-9B-Thinking~\citep{Hong2025GLM41VThinkingTV}; Appendix~\ref{app:data_construction} describes the construction and filtering procedure. The default backbone is ``\texttt{Qwen3-VL-8B-Instruct}''~\citep{Bai2025Qwen3VLTR}. We freeze the vision encoder and fine-tune the LLM and projector; all variants within each comparison share the same optimization schedule and single-vector evaluation protocol. We train each variant for one epoch and report its final checkpoint. During training, Boundary Readout operates on teacher-forced semantic-evidence target sequences. During inference, the model generates the evidence sequence with greedy decoding and constructs readouts at the designated positions. The readouts are averaged and L2-normalized to obtain the final embedding. Detailed hyperparameter settings are listed in Appendix~\ref{app:hyperparameter_config}.

\paragraph{Baselines.}
We compare EviAlign with three families of MLLM-based multimodal embedders. Direct embedding baselines include GME~\citep{Zhang2024GMEIU}, VLM2Vec~\citep{Jiang2024VLM2VecTV}, VLM2Vec-V2~\citep{Meng2025VLM2VecV2AM}, UniME-V2~\citep{Gu2025UniMEV2MF}, and LamRA~\citep{Liu2024LamRALM}. Explicit generation or reasoning baselines include UME-R1~\citep{Lan2025UMER1ER}, RIME~\citep{Wu2026RIME}, and Think-Then-Embed~\citep{Cui2025ThinkTE}; latent-reasoning baselines include PLUME~\citep{He2026PLUME} and LaME~\citep{Wu2026LaME}.

Appendix~\ref{app:baseline_details} summarizes each baseline's reported training scale and representation extraction.

\subsection{Main Results}\label{exp:main_results}
Table~\ref{tab:main_results} reports final retrieval performance and the backbone used by each method. With 500K query--candidate training pairs, EviAlign reaches 75.5 average Recall@1 with Qwen2-VL-7B and 76.9 with Qwen3-VL-8B. Under the Qwen2-VL-7B backbone, EviAlign exceeds UME-R1 and RIME. Think-Then-Embed$_t$ reports 75.9 using a Qwen2-VL-7B embedder together with a separate Qwen2.5-VL-72B reasoner. The Qwen3-VL-8B model reaches 81.6 in-domain and 67.7 out-of-domain, while GME records the strongest out-of-domain average at 71.8. The controlled studies below use Qwen3-VL-8B and hold the data budget, optimization, and evaluation protocol fixed across variants. Figure~\ref{fig:backbone_ablation} shows scaling within EviAlign.

\subsection{Controlled Study of Evidence--Readout Co-design}
\label{sec:ablation}

\paragraph{Factorial Evidence--Readout Comparison.}
We vary evidence organization and representation readout in a $2\times3$ study under the same 500K budget. For each training example, the Semantic and Mixed targets contain the same five labeled evidence spans. Semantic maintains a fixed correspondence between evidence roles and boundary positions; Mixed permutes the labeled spans for each example while keeping the boundary-token identities and order fixed. The readout factor compares (i) \emph{Trailing}, where only the final-layer state at \texttt{<SUM>} serves as the embedding for contrastive training and retrieval; (ii) \emph{Distributed}, which inserts five distinct readout tokens at positions $\bar p_k=\lfloor kT/5\rfloor$ for $k=1,\ldots,5$, where $T$ is the evidence-target length before insertion; and (iii) \emph{Boundary}, which places the same five token types at the five evidence-unit ends. Both five-readout conditions mean-pool the token states. Trailing provides the single-state reference; Distributed matches Boundary in readout count and pooling while placing those states independently of evidence boundaries. All six configurations are trained independently under the same data budget and optimization schedule. At inference, each model generates its own sequence and constructs the embedding from the resulting readout states. Appendix~\ref{app:factorial_construction} gives the complete construction.

Let $\operatorname{R@1}(o,r)$ denote average Recall@1 for evidence organization $o\in\{M,S\}$ and readout $r\in\{T,D,B\}$, corresponding to mixed or semantic evidence and trailing, distributed, or boundary readout. We define the organization advantage under readout $r$ as $A_r=\operatorname{R@1}(S,r)-\operatorname{R@1}(M,r)$. Using the distributed condition as the matched five-readout control, the co-design interaction is
\begin{equation}
    \Delta_{\mathrm{co}}
    = A_B-A_D
    = \bigl[\operatorname{R@1}(S,B)-\operatorname{R@1}(M,B)\bigr]
    - \bigl[\operatorname{R@1}(S,D)-\operatorname{R@1}(M,D)\bigr].
    \label{eq:codesign_interaction}
\end{equation}

\begin{table}[htbp]
    \centering
    \caption{Evidence organization $\times$ readout on MMEB (average Recall@1, \%).}
    \label{tab:evidence_readout_factorial}
    \setlength{\tabcolsep}{7pt}
    \renewcommand{\arraystretch}{1.14}
    \begin{tabularx}{\linewidth}{l*{4}{>{\centering\arraybackslash}X}}
        \toprule
        \multirow{2}{*}{\makecell[l]{\textbf{Evidence}\\\textbf{organization}}} & \multicolumn{3}{c}{\textbf{Readout}} & \multirow{2}{*}{\makecell{\textit{Boundary} $-$\\\textit{Distributed}}} \\
        \cmidrule(lr){2-4}
        & \textbf{Trailing} & \textbf{Distributed} & \textbf{Boundary} & \\
        \midrule
        \rowcolor{EviCoralTint} Mixed & 73.60 & 74.20 & 74.55 & +0.35 \\
        \rowcolor{EviBlueTint} Semantic & 73.74 & 74.85 & \textbf{76.94} & +2.09 \\
        \midrule
        \textit{Semantic $-$ Mixed} & +0.14 & +0.65 & \textbf{+2.39} & \textbf{+1.74} \\
        \bottomrule
    \end{tabularx}
\end{table}

Table~\ref{tab:evidence_readout_factorial} reports the six results visualized in Figure~\ref{fig:intro_challenge}(b). Its bottom row compares Semantic with Mixed under each readout, while its rightmost column compares Boundary with Distributed under each evidence organization. Under Distributed, the Semantic--Mixed advantage is 0.65 points; with Boundary, it grows to 2.39. Viewed from the other direction, moving the readouts from distributed positions to boundaries adds 0.35 points for Mixed but 2.09 for Semantic. Both comparisons give the same co-design interaction, $\Delta_{\mathrm{co}}=2.39-0.65=2.09-0.35=1.74$ points. Across the six end-to-end configurations, stable role-to-boundary correspondence produces its largest Semantic--Mixed advantage when the model constructs the embedding from the emitted boundary states.

In EviAlign, the boundary states form the interface between generated evidence and the retrieval embedding. Each state is read after its evidence unit has been completed, with the preceding units available through causal context. Contrastive learning trains the pooled boundary states as the retrieval representation, and inference applies the same readout rule to the model's generated evidence. Distributed preserves the token inventory and aggregation without aligning readouts to evidence-unit ends; Mixed preserves the labeled evidence spans without consistent role-to-boundary assignments. The evidence schema thus specifies both what the model generates and which states become the vector used for ranking.

\Needspace{6\baselineskip}
\subsection{Analysis of Representation Construction}
\label{sec:representation_analysis}

\begin{figure}[b]
    \centering
    \begin{subfigure}[c]{0.54\linewidth}
        \centering
        \includegraphics[width=\linewidth]{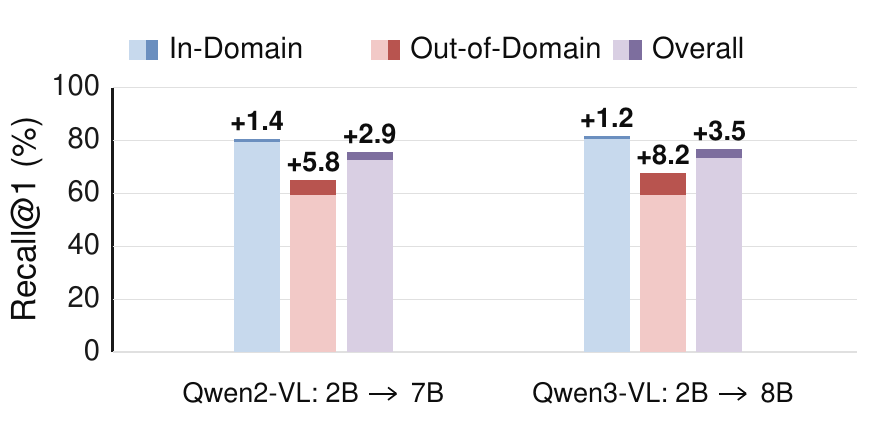}
        \caption{Backbone scaling.}
        \label{fig:backbone_ablation}
    \end{subfigure}\hfill
    \begin{subfigure}[c]{0.44\linewidth}
        \centering
        \includegraphics[width=\linewidth]{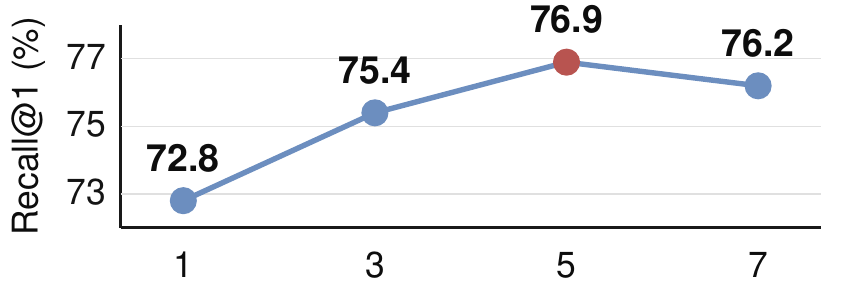}
        \caption{Evidence-unit count.}
        \label{fig:slot_count}
    \end{subfigure}
    \caption{Analysis of EviAlign's representation construction under the 500K training budget. (a) Performance across Qwen-VL model generations and scales; darker regions show the absolute gains from the larger model within each generation. (b) Effect of the joint evidence-unit and boundary-readout configuration on MMEB Recall@1.}
\end{figure}

\paragraph{Generation, Readout Capacity, and Evidence Source.}
\begin{table}[htbp]
    \centering
    \caption{Controlled decomposition of generation and representation readout on MMEB (average Recall@1, \%).}
    \label{tab:generation_teacher_controls}
    \small
    \setlength{\tabcolsep}{5pt}
    \renewcommand{\arraystretch}{1.12}
    \begin{adjustbox}{max width=\columnwidth}
    \begin{tabular}{lllr}
        \toprule
        \textbf{Evidence Generation} & \textbf{Representation Readout} & \textbf{Teacher} & \textbf{Overall} \\
        \midrule
        None & Single trailing & -- & 68.30 \\
        None & Multiple readouts & -- & 70.30 \\
        \midrule
        \rowcolor{EviCoralTint} Free-form CoT & Single trailing & GLM & 73.75 \\
        \rowcolor{EviCoralTint} Semantic evidence & Single trailing & GLM & 73.74 \\
        \rowcolor{EviCoralTint} Semantic evidence & Multiple trailing readouts (MLTR) & GLM & 73.07 \\
        \rowcolor{EviCoralTint} Free-form CoT + semantic evidence & Boundary & GLM & 76.59 \\
        \midrule
        \rowcolor{EviBlueTint} Semantic evidence & Boundary & Qwen & 75.60 \\
        \rowcolor{EviBlueTint} Semantic evidence & Boundary & GLM & \textbf{76.94} \\
        \bottomrule
    \end{tabular}
    \end{adjustbox}
    \vspace{2pt}
    \begin{minipage}{0.96\columnwidth}
    \footnotesize All variants use the same Qwen3-VL-8B student, 500K training pairs, and optimization settings. GLM and Qwen denote GLM-4.1V-9B-Thinking and Qwen3-VL-8B-Thinking. MLTR appends five dedicated readout tokens to the generated sequence end.
    \end{minipage}
\end{table}

Table~\ref{tab:generation_teacher_controls} decomposes the contributions of generation, readout capacity, and evidence source under the same 500K training budget. Adding free-form CoT before a trailing readout raises average Recall@1 from 68.30 to 73.75. With the same trailing readout, semantic evidence and free-form CoT are nearly tied (73.74 vs. 73.75).

Readout multiplicity is likewise insufficient by itself: five readouts without generation reach 70.30, while five trailing readouts after semantic evidence reach 73.07. Using the same semantic-evidence supervision, the boundary-readout model reaches 76.94, outperforming the separately trained trailing-readout model by 3.20 points. Prepending free-form CoT to the semantic evidence gives 76.59 with the same boundary readout, while changing the evidence teacher from GLM-4.1V-9B-Thinking~\citep{Hong2025GLM41VThinkingTV} to Qwen3-VL-8B-Thinking~\citep{Bai2025Qwen3VLTR} gives 75.60. Together, these results show that structured evidence realizes its retrieval advantage when its semantic units also determine where representations are read.

\paragraph{Representation-Construction Cost and Format.} At batch size 4, end-to-end batch times are 94.6 ms for direct encoding, 1,502.5 ms for EviAlign, and 9,825.3 ms for CoT-plus-semantic Boundary Readout. The latter two average 114.3 and 702.1 generated tokens and reach 76.94 and 76.59 Recall@1, respectively: the longer CoT prefix increases time without improving retrieval. Of 12,000 query generations, 99.52\% contain all five boundary tokens exactly once and in order. Appendices~\ref{app:boundary_format_validity} and~\ref{app:complexity} report per-task format validity and full efficiency results.

\paragraph{Backbone Analysis.} Figure~\ref{fig:backbone_ablation} shows that scaling from 2B to 7B/8B improves the overall result from 72.6 to 75.5 for Qwen2-VL and from 73.4 to 76.9 for Qwen3-VL. Out-of-domain gains are 5.8 and 8.2 points, respectively, compared with in-domain gains of 1.4 and 1.2 points. EviAlign improves with backbone scale in both model families, with the largest gains on the out-of-domain tasks. Appendix Table~\ref{tab:backbone_per_task_500k} gives the per-task 7B/8B results.

\paragraph{Effect of Evidence-Unit Configuration.}
We train $K \in \{1,3,5,7\}$ evidence-schema and readout configurations under identical settings. They respectively use Entity; Entity/Attribute/Relation; the default five units; and two additional units, Context (scene context) and Emotion (emotional tone). Figure~\ref{fig:slot_count} rises from 72.8 ($K=1$) to 75.4 ($K=3$) and 76.9 ($K=5$), then falls to 76.2 ($K=7$). The five-unit schema performs best among these joint configurations; adding further units does not necessarily improve retrieval. Appendix~\ref{app:prompt} gives the prompts.

\paragraph{Readout Complementarity.} Without retraining, we evaluate individual boundary readouts and leave-one-out aggregates from the full five-unit model. These are contextualized states rather than isolated semantic features: each has access to the input and the evidence generated up to its boundary. The one-slot and leave-one-out evaluations therefore offer complementary views of a boundary's retrieval information and its contribution to the pooled vector. ENT alone reaches 76.12 average Recall@1, while aggregating all five readouts reaches 76.94; removing any one reduces the average by 0.23--0.36 points. Different tasks favor different individual readouts: Detail is the strongest single readout on CIRR (78.6), Attribute on VisualNews image-to-text (84.8), and Relation on OVEN (70.9). On OVEN, the full aggregate reaches 72.2. Full aggregation exceeds every individual readout on seven tasks and yields the highest 12-task average. Fixed aggregation thus combines cues across tasks without selecting a task-specific readout. These inference-time interventions differ from the $K$-unit comparisons in Figure~\ref{fig:slot_count}, whose models are trained separately. Appendices~\ref{app:leave_one_out} and~\ref{app:readout_geometry} report per-task and geometry results; Appendices~\ref{app:output_discriminability}--\ref{app:complexity} and~\ref{app:case_studies} provide the remaining analyses and examples.

\begin{figure}[H]
    \centering
    \begin{tcolorbox}[
        enhanced,
        colframe=EviBlue,
        colback=white,
        boxrule=0.65pt,
        arc=2pt,
        left=4pt,right=4pt,top=4pt,bottom=4pt
    ]
    \begin{minipage}[c]{0.35\linewidth}
        \begin{minipage}[c]{0.43\linewidth}
            \centering
            \parbox[c][1.55cm][c]{\linewidth}{\centering\includegraphics[width=\linewidth,height=1.55cm,keepaspectratio]{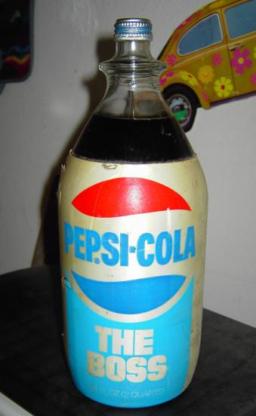}}\par
            \vspace{2pt}{\scriptsize\textbf{Query image}}
        \end{minipage}\hfill
        \begin{minipage}[c]{0.53\linewidth}
            \centering
            \parbox[c][1.55cm][c]{\linewidth}{\centering\scriptsize\itshape ``Show three bottles of soft drink.''}\par
            \vspace{2pt}{\scriptsize\textbf{Query text}}
        \end{minipage}
    \end{minipage}\hfill
    \begin{minipage}[c]{0.15\linewidth}
        \centering
        \parbox[c][1.55cm][c]{\linewidth}{\centering\includegraphics[width=\linewidth,height=1.55cm,keepaspectratio]{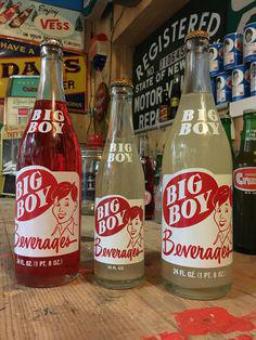}}\par
        \vspace{2pt}{\scriptsize\textbf{Retrieved image}}
    \end{minipage}\hfill
    \begin{minipage}[c]{0.46\linewidth}
        \scriptsize
        \textbf{Entity:} three vintage soft-drink glass bottles with crown caps\\
        \textbf{Attribute:} warm indoor, retro product-photography setting\\
        \textbf{Relation:} three bottles standing upright side by side\\
        \textbf{Detail:} clear glass, dark liquid, printed labels, metal caps\\
        \textbf{Summary:} three vintage soft-drink bottles arranged upright side by side
    \end{minipage}
    \end{tcolorbox}
    \vspace{-3pt}
    \caption{A CIRR example. The query pairs a reference image with a text modification; the generated evidence describes the requested three-bottle target while retaining visual attributes of the reference.}
    \label{fig:main_cirr_case}
\end{figure}

\paragraph{Qualitative Example.} Figure~\ref{fig:main_cirr_case} pairs a reference image of one bottle with a text request for three. Entity records the requested count, Relation specifies the side-by-side arrangement, and Detail carries over visual cues from the reference bottle. Summary combines the preserved appearance with the new count and arrangement. The resulting evidence describes the intended retrieval target rather than the observed image alone, and its boundary states are pooled into the embedding used for ranking. Appendix~\ref{app:case_studies} provides examples from two additional retrieval settings.

\Needspace{13\baselineskip}
\section{Conclusion}
We introduced \textbf{EviAlign}, which uses the same evidence boundaries to organize generation and construct a multimodal retrieval embedding. Its five contextualized boundary readouts are aggregated into one vector for standard indexing and scoring, so generation changes representation construction without changing the retrieval index. A controlled $2\times3$ study shows that semantic evidence yields its largest advantage when readouts follow the corresponding boundaries, producing a 1.74-point co-design interaction. Complementary controls show that semantic evidence and free-form CoT are nearly tied under a trailing readout, while additional readout states alone do not reproduce the boundary result. With 500K training pairs, EviAlign achieves 76.9 average Recall@1 across 12 MMEB retrieval tasks. Together, these results support using semantic evidence structure not only to organize generation, but also to determine where retrieval representations are read.

\subsection*{Reproducibility Statement}
Section~\ref{sec:exp_setup} specifies the MMEB retrieval evaluation protocol. Appendix~\ref{app:algorithm} provides the training and inference algorithm; Appendix~\ref{app:experimental_details} documents data construction, filtering criteria, hyperparameters, and the controlled experimental configurations. Appendix~\ref{app:prompt} presents the evidence-generation, annotation, evidence-configuration ablation, and judge prompt templates. Per-task results and additional analyses are reported in Appendix~\ref{app:leave_one_out}--\ref{app:case_studies}.

\subsection*{AI Use Statement}
We used GLM-4.1V-9B-Thinking and Qwen3-VL-8B-Thinking to generate structured evidence annotations for the training experiments, and GPT-4o to assess training-target quality. Generative AI tools also assisted with language polishing. The authors reviewed AI-assisted content and take responsibility for the final paper.

\clearpage
\appendix
\raggedbottom

\section{EviAlign Algorithm}
\label{app:algorithm}
Algorithm~\ref{algo:evialign} summarizes the training and inference procedure of EviAlign.
\RestyleAlgo{ruled}
\begin{algorithm}
\small
    \SetAlgoNlRelativeSize{-1}
    \SetAlgoLined
    \caption{EviAlign Training and Inference}\label{algo:evialign}
    \SetKwInOut{Input}{Input}
    \SetKwInOut{Output}{Output}
    \SetKwFunction{Embed}{GetEmbedding}

    Initialize MLLM $\mathcal{M}$.

    \LinesNumbered
   
    \SetKwProg{Fn}{Function}{:}{}
    \Fn{\Embed{$x, \mathbf{y}$ (optional)}}{
        Use target evidence $\mathbf{y}$ under teacher forcing during training; otherwise generate $\mathbf{z}$; \tcp{Eq. \ref{eq:structured_evidence}}
        Read boundary representations $\{\mathbf{e}_k\}_{k=1}^K$; \tcp{Eq. \ref{eq:evidence_slot_readout}}
        Mean-pool and normalize the readouts to obtain $\mathbf{h}$; \tcp{Eq. \ref{eq:evidence_aggregation}}
        \KwRet $\mathbf{h}, \{\mathbf{e}_k\}_{k=1}^K$
    }
    
    \BlankLine
    
    \If{is training}{
        \Input{Batch of query--candidate pairs and target evidence sequences.}
        \ForEach{pair $(q_n, c_n) \in \mathcal{B}$}{
            $\mathbf{h}_{q_n}, \{\mathbf{e}^{(q)}_k\}_{k=1}^K \leftarrow$ \Embed{$q_n, \mathbf{y}_{q_n}$}; \\
            $\mathbf{h}_{c_n}, \{\mathbf{e}^{(c)}_k\}_{k=1}^K \leftarrow$ \Embed{$c_n, \mathbf{y}_{c_n}$};
        }
        
        Compute contrastive loss $\mathcal{L}_{\text{NCE}}$ over $\{\mathbf{h}_{q_n}, \mathbf{h}_{c_n}\}_{n=1}^{|\mathcal{B}|}$; \tcp{Eq. \ref{eq:contrastive_loss}}
        Compute language modeling loss $\mathcal{L}_{\text{LM}}$ over semantic-evidence sequences; \tcp{Eq. \ref{eq:lm_loss}}
        Compute total loss $\mathcal{L} \leftarrow \mathcal{L}_{\text{NCE}} + \mathcal{L}_{\text{LM}}$; \\
        Update MLLM parameters.
    }
    
    \BlankLine
    
    \If{is inference}{
        \Input{Query $q$, candidate corpus $\mathcal{C}$, and precomputed candidate embeddings $\{\mathbf{h}_c:c\in\mathcal{C}\}$.}
        $\mathbf{h}_q, \_ \leftarrow$ \Embed{$q, \varnothing$}; \\
        Retrieve top-$k$ candidates $\mathcal{C}^{\star}$ using $\mathbf{h}_q^{\top}\mathbf{h}_c$; \tcp{Eq. \ref{eq:retrieval_similarity}}
        \KwRet $\mathcal{C}^{\star}$
    }

\end{algorithm}

\section{Experimental Details}
\label{app:experimental_details}

\subsection{Benchmark Details}
\label{app:benchmarks}
We provide a comprehensive overview of the 12 benchmarks that constitute the MMEB retrieval evaluation suite.
These datasets collectively span a broad spectrum of multimodal retrieval challenges---from fundamental image-text alignment and composed image retrieval to knowledge-intensive entity grounding, fine-grained visual attribute matching, and visual document understanding.
A summary of each benchmark's modality configurations and domain categorization is presented in Table~\ref{tab:mmeb_summary}; detailed per-dataset descriptions follow.

\subsubsection{In-Domain Datasets}

\paragraph{VisDial (Visual Dialogue)~\citep{das2017visual}.} Originating from a conversational AI challenge, the dataset features dialogues between a ``Questioner'' (who sees only a caption) and an ``Answerer'' (who sees the image). In the MMEB benchmark, this dataset is adapted into a discriminative retrieval task. Instead of generating responses, the model is evaluated on its ability to retrieve the correct target image given the full dialogue history, effectively testing its ability to resolve visual references within multi-turn conversational contexts.

\paragraph{CIRR (Composed Image Retrieval on Real-life images)~\citep{liu2021image}.} The dataset is a standard benchmark for Composed Image Retrieval (CIR) containing open-domain, real-life photography. Each query consists of a reference image and a relative modification text (e.g., ``change the dog to a cat''). The task requires the model to retrieve a target image that visually reflects the textual instructions applied to the reference. This setup challenges the model to understand fine-grained visual semantics and compositional logic beyond simple keyword matching.

\paragraph{VisualNews~\citep{liu2021visual}.} The dataset is a large-scale entity-aware dataset consisting of news photographs paired with their original captions. The MMEB suite evaluates this dataset under two distinct bidirectional setups: \textbf{VisualNews t2i} (retrieving the relevant image given a caption) and \textbf{VisualNews i2t} (retrieving the correct caption given an image). These tasks assess the model's capability to ground complex, event-driven textual concepts---often rich in named entities and context---into the visual domain.

\paragraph{MSCOCO (Microsoft Common Objects in Context)~\citep{lin2014microsoft}.} As a cornerstone benchmark in vision-language research, the dataset contains high-quality images of everyday scenes annotated with five human-written captions. We evaluate on the retrieval splits defined by the MMEB benchmark, covering both \textbf{MSCOCO t2i} and \textbf{MSCOCO i2t}. Performance on this dataset serves as a baseline for the model's fundamental ability to perform semantic alignment in general visual scenarios with complex spatial layouts.

\paragraph{NIGHTS~\citep{fu2023dreamsim}.} The dataset focuses on visual similarity across diverse environmental conditions. The original dataset contains triplets with human judgments on similarity. Following the protocol established in M-BEIR~\citep{Wei2024UniIRTA} and adopted by MMEB, this is formulated as a pairwise retrieval task. The reference image serves as the query, and the specific perturbed version identified by human annotators as the ``match'' serves as the ground-truth target, evaluating robustness to domain shifts.

\paragraph{WebQA~\citep{chang2022webqa}.} The dataset is originally a multihop, multimodal question-answering dataset mimicking web search behavior. For the retrieval evaluation, MMEB isolates the evidence retrieval stage, where the goal is to select the correct candidate that contains the answer to a natural language query. This formulation tests the model's capacity for knowledge-intensive retrieval and multihop reasoning.

\subsubsection{Out-of-Domain Datasets}
\paragraph{FashionIQ~\citep{wu2021fashion}.} As a domain-specific counterpart to CIRR, the dataset focuses on the fashion industry (Dresses, Shirts, Tops). It uses crowd-sourced relative captions describing specific attribute differences between a reference product and a target product (e.g., ``is darker blue and has a V-neck''). This benchmark specifically tests the model's ability to comprehend fine-grained attribute manipulation and visual feedback within a specialized vertical domain.

\paragraph{Wiki-SS-NQ (Wikipedia ScreenShot Natural Questions)~\citep{Ma2024UnifyingMR}.} Derived from the Natural Questions (NQ) dataset~\citep{Kwiatkowski2019NaturalQA}, the dataset introduces a ``visual document understanding'' component. Instead of retrieving plain text, the task involves retrieving screenshots of Wikipedia pages (Wiki-SS) that answer a user's question. This unique format requires the model to process not just textual content but also structural and layout information embedded in the page screenshot.

\paragraph{OVEN (Open-domain Visual Entity Recognition)~\citep{hu2023open}.} The dataset evaluates entity grounding in an open-world setting. Each instance pairs a visual query with a recognition-based question. The retrieval task defined in MMEB involves linking this query to a specific knowledge entry. The candidate pool consists of Wikipedia images accompanied by their textual descriptions (titles and summaries). Models must effectively bridge visual features with encyclopedic textual knowledge to identify the correct entity within the MMEB candidate pool.

\paragraph{EDIS (Entity-Driven Image Search)~\citep{Liu2023EDISEI}.} The dataset addresses cross-modal retrieval in the dynamic news domain. Unlike generic caption retrieval, EDIS queries are typically rich in specific named entities and event descriptions. The evaluation setup matches these queries against a pool of news images paired with their headlines. This requires the model to go beyond surface-level matching and perform deep semantic reasoning to link textual entities with their visual representations.

\begin{table}[t]
\centering
\caption{Summary of the MMEB retrieval evaluation suite. ``Query'' and ``Target'' denote input modalities: T~=~text, I~=~image, I+T~=~image-text pair.}
\label{tab:mmeb_summary}
\small
\setlength{\tabcolsep}{3pt}
\renewcommand{\arraystretch}{1.15}
\begin{tabularx}{\linewidth}{l l c c X}
\toprule
\textbf{Split} & \textbf{Dataset} & \textbf{Query} & \textbf{Target} & \textbf{Retrieval Task} \\
\midrule
\multirow{8}{*}{{\textit{In Domain}}}
& VisDial          & T     & I     & Multi-turn dialogue resolution \\
& CIRR             & I+T   & I     & Composed image retrieval \\
& VisualNews (t2i) & T     & I     & Entity-grounded news matching \\
& VisualNews (i2t) & I     & T     & Visual-to-caption grounding \\
& MSCOCO (t2i)     & T     & I     & General visual-language alignment \\
& MSCOCO (i2t)     & I     & T     & General visual-language alignment \\
& NIGHTS           & I     & I     & Fine-grained visual similarity \\
& WebQA            & T     & I+T   & Knowledge-intensive multi-hop QA \\
\midrule
\multirow{4}{*}{{\textit{Out of Domain}}}
& OVEN             & I+T   & I+T   & Open-domain visual entity recognition \\
& FashionIQ        & I+T   & I     & Fine-grained attribute manipulation \\
& EDIS             & T     & I+T   & Named-entity news image search \\
& Wiki-SS-NQ       & T     & I     & Visual document understanding \\
\bottomrule
\end{tabularx}
\end{table}

\subsection{Baseline Details}
\label{app:baseline_details}
We provide detailed specifications for the baselines compared in our experiments, focusing on differences in training data scale, representation construction, and retrieval protocol that contextualize comparisons with EviAlign.

\paragraph{GME~\citep{Zhang2024GMEIU}.}
GME is a data-intensive approach built on the Qwen2-VL~\citep{Wang2024Qwen2VLEV} backbone. To achieve broad coverage across arbitrary modality combinations, it relies on a composite dataset of approximately 8 million samples, integrating diverse open-source benchmarks and $\sim$1.1M self-synthesized fused-modal data. Embeddings are extracted from the hidden state of the last (EOS) token using standard instruction-based prompting.

\paragraph{LamRA-Ret~\citep{Liu2024LamRALM}.}
Table~\ref{tab:main_results} uses the released \href{https://huggingface.co/code-kunkun/LamRA-Ret}{LamRA-Ret checkpoint} based on Qwen2-VL-7B. Its in-domain, out-of-domain, and overall retrieval means are computed from the corresponding per-task results in the \href{https://huggingface.co/spaces/TIGER-Lab/MMEB-Leaderboard/blob/026e4b2c1e45dee2b01cb037c7e8114908576414/scores/LamRA-Ret.json}{public MMEB leaderboard}.
LamRA empowers MLLMs with retrieval capabilities through a multi-stage adaptation process using lightweight LoRA modules. Its training pipeline involves a total of approximately 1.4 million samples, combining a language-only pre-training stage on the NLI dataset (275k) with a multimodal instruction tuning stage on M-BEIR~\citep{Wei2024UniIRTA} (1.1M). For embedding extraction, the model utilizes specific summarization prompts (``Summarize above image in one word''), and the representation is obtained from the hidden state immediately preceding a specific placeholder token.

\paragraph{VLM2Vec~\citep{Jiang2024VLM2VecTV} \& VLM2Vec-V2~\citep{Meng2025VLM2VecV2AM}.}
These models represent the standard contrastive fine-tuning paradigm for MLLMs. VLM2Vec (V1) uses a balanced 662k-sample subset of MMEB. V2 significantly expands the corpus to 1.7 million samples by incorporating video instruction data, visual documents, and the original image data, with an interleaved sub-batching strategy to handle modality heterogeneity. Both extract embeddings from the last token's hidden state.

\paragraph{UniME-V2~\citep{Gu2025UniMEV2MF}.}
UniME-V2 trains on 662K pairs from the 20 in-domain MMEB datasets and uses MLLM judgments for hard-negative mining and soft semantic supervision. Its Qwen2-VL-7B encoder reports 73.1 average Precision@1 over MMEB's 12 retrieval tasks.

\paragraph{UME-R1~\citep{Lan2025UMER1ER}.}
UME-R1 introduces a two-stage reasoning-driven paradigm. The Cold-Start SFT stage uses 1.5 million samples augmented with free-form CoT reasoning traces and summaries from a thinking model. A subsequent RL stage (11k balanced samples across modalities) further optimizes the reasoning process via GRPO. Embeddings are extracted from a dedicated generative token appended after the reasoning sequence, bridging generation and retrieval.

\paragraph{RIME~\citep{Wu2026RIME}.}
RIME replaces free-form chain-of-thought with a retrieval-oriented rewrite and jointly optimizes generation and embedding through Cross-Mode Alignment and Refine-RL.

\paragraph{Think-Then-Embed~\citep{Cui2025ThinkTE}.}
Think-Then-Embed first generates embedding-oriented context with a reasoner and then constructs the retrieval representation with an embedder conditioned on both the input and generated context. The reported TTE$_t$ variant uses a separate teacher reasoner.

\paragraph{PLUME~\citep{He2026PLUME}.}
PLUME replaces explicit textual reasoning with a short autoregressive rollout of continuous latent states and transfers explicit reasoning behavior into latent computation through a progressive curriculum.

\paragraph{LaME~\citep{Wu2026LaME}.}
LaME performs embedding-oriented reasoning with a fixed number of learnable reason tokens in a single forward pass, using them as a fixed-capacity latent bottleneck.

\subsection{EviAlign-Evidence Construction}
\label{app:data_construction}
To teach the MLLM to directly generate semantic retrieval evidence, we construct \textbf{EviAlign-Evidence}, a supervised training corpus of $\langle$multimodal input, semantic-evidence target$\rangle$ pairs from the MMEB-V1 retrieval training split.
Specifically, we employ ``GLM-4.1V-9B-Thinking''~\citep{Hong2025GLM41VThinkingTV} to convert each training input from the MMEB-V1~\citep{Jiang2024VLM2VecTV} retrieval subset into the semantic-evidence format consumed by EviAlign. The four out-of-domain evaluation datasets are excluded from EviAlign-Evidence construction. Given an input, a specialized prompt organizes task-relevant evidence into explicit semantic units and appends the corresponding boundary tokens (see Appendix~\ref{app:prompt} for the complete template).
To ensure annotation quality and training stability, we apply automatic filtering to remove samples exhibiting excessive repetition, invalid formats, or evidence sequences exceeding the 8192-token hard safety limit. After filtering, the retained semantic evidence is combined with the corresponding boundary tokens and aligned back to the original MMEB retrieval examples using their source identifiers. This process retains approximately 500K query--candidate training pairs. Their evidence targets supervise Semantic Evidence Generation and provide the teacher-forced sequences used by Boundary Readout during training.

\subsection{Hyperparameter Configuration}
\label{app:hyperparameter_config}
To facilitate reproducibility, we detail the hyperparameter settings and hardware configurations used for EviAlign in Table~\ref{tab:hyperparams}.

\begin{table}[ht] 
    \centering      
    \caption{Hyperparameter settings and hardware configurations for EviAlign.}
    \label{tab:hyperparams} 

    \begin{tabularx}{\linewidth}{l X}
    \toprule
    \textbf{Hyperparameter} & \textbf{Value} \\
    \midrule
    \multicolumn{2}{l}{\textit{Model Architecture}} \\
    \hspace{1.5em}Base Backbone & Qwen3-VL-8B-Instruct \\
    \hspace{1.5em}Vision Encoder & Frozen \\
    \hspace{1.5em}LLM \& Projector & Full Fine-Tuning \\
    \hspace{1.5em}Boundary Token Initialization & \texttt{[EOS]} embedding \\
    \hspace{1.5em}Min / Max Pixels & $768$ / $1,572,864$ \\
    \hspace{1.5em}Max Sequence Length & 12,288 tokens \\
    \hspace{1.5em}Precision & BF16 \\
    \midrule
    \multicolumn{2}{l}{\textit{Optimization}} \\
    \hspace{1.5em}Optimizer & AdamW \\
    \hspace{1.5em}Learning Rate & $5 \times 10^{-5}$ \\
    \hspace{1.5em}Contrastive Temperature $\tau$ & $0.02$ \\
    \hspace{1.5em}LM Loss Reduction & Per-side mean over supervised target tokens \\
    \hspace{1.5em}NCE / Query LM / Candidate LM Weights & $1.0$ / $1.0$ / $1.0$ \\
    \hspace{1.5em}LR Schedule & Cosine Decay \\
    \hspace{1.5em}Warmup Ratio & 0.03 \\
    \hspace{1.5em}Weight Decay & 0 \\
    \hspace{1.5em}Gradient Clipping & 1.0 \\
    \hspace{1.5em}DeepSpeed Strategy & ZeRO Stage 3 \\
    \midrule
    \multicolumn{2}{l}{\textit{Training Setup}} \\
    \hspace{1.5em}Batch Size per GPU & 4 pairs \\
    \hspace{1.5em}Gradient Accumulation Steps & 2 \\
    \hspace{1.5em}Effective Global Batch Size & 256 pairs \\
    \hspace{1.5em}Training Epoch & 1 \\
    \hspace{1.5em}Random Seed & 42 \\
    \hspace{1.5em}Hardware & 32 $\times$ NVIDIA H20 (80GB) \\
    \bottomrule
    \end{tabularx}
\end{table} 
\subsection{Generation, Readout-Capacity, and Evidence-Teacher Controls}
\label{app:generation_teacher_controls}
Table~\ref{tab:generation_teacher_controls} reports the controlled generation and readout ablations discussed in Section~\ref{sec:representation_analysis}. The no-generation multi-readout control uses five dedicated readout tokens in place of a generated evidence sequence. All five tokens participate in training, and their final-layer states are mean-pooled and L2-normalized. MLTR places the five dedicated readout tokens together after the generated semantic evidence. The free-form-CoT trailing control generates an unstructured reasoning sequence and uses a single trailing readout. The CoT-plus-semantic control generates free-form reasoning followed by structured semantic evidence and retains the five boundary readouts. Both CoT controls use GLM-generated targets, as specified in Table~\ref{tab:generation_teacher_controls}.

\subsection{Factorial Evidence--Readout Construction}
\label{app:factorial_construction}

The $2\times3$ study in Section~\ref{sec:ablation} matches evidence content in the training targets while varying cross-example role consistency and readout location. In the \emph{semantic} condition, Entity, Attribute, Relation, Detail, and Summary evidence maintain fixed assignments to \texttt{<ENT>}, \texttt{<ATT>}, \texttt{<REL>}, \texttt{<DET>}, and \texttt{<SUM>}. In the \emph{mixed} condition, one sample-specific permutation moves the same five evidence spans together with their textual role labels across these fixed boundary-token positions. The permutation is sampled once during target construction and retained across training epochs. At inference, each separately trained model autoregressively generates its own five-unit sequence, and representations are collected from the emitted token positions according to the assigned readout condition. The training targets therefore preserve the labeled evidence spans, unit count, boundary-token identities and order, and length distribution while varying the cross-example role-to-boundary correspondence.

For representation readout, the \emph{trailing} condition retains all five boundary tokens but uses only the final-layer state at \texttt{<SUM>} for contrastive training and retrieval. The \emph{distributed} condition places five dedicated readout tokens at content-independent positions
\begin{equation}
    \bar p_k=\left\lfloor\frac{kT}{5}\right\rfloor,
    \qquad k=1,\ldots,5,
    \label{eq:distributed_readout_positions}
\end{equation}
where $T$ is the complete evidence-sequence length before readout-token insertion. During supervised-target construction, $T$ is known and the five readout tokens are inserted at these positions before training; the first four are evenly spaced through the sequence and the fifth follows the complete sequence. The model learns to generate the readout tokens jointly with the evidence sequence; at inference, their hidden states are collected when the model autoregressively emits the corresponding tokens in the same generation pass. Their target placement is determined only by sequence length and is not selected or adjusted using evidence boundaries. The \emph{boundary} condition instead places the same five distinct token types---\texttt{<ENT>}, \texttt{<ATT>}, \texttt{<REL>}, \texttt{<DET>}, and \texttt{<SUM>}---at the five evidence-unit ends. The two five-readout conditions share token identities, order, count, initialization, training treatment, and mean aggregation followed by L2 normalization; they differ in the rule used to place the readout tokens in the training targets. All six configurations retain the same five evidence units and are trained separately using the same Qwen3-VL-8B-Instruct backbone, 500K training pairs, optimization schedule, and greedy inference procedure. The comparison tests whether stable semantic evidence organization is most useful when representation readout follows the corresponding organization.

\begin{table}[t]
    \centering
    \caption{Schematic of the $2\times3$ evidence order and readout for a CIRR query requesting cookies on a counter. The five evidence spans are abbreviated, and Mixed shows one example permutation.}
    \label{tab:factorial_construction_example}
    \normalsize
    \setlength{\tabcolsep}{5pt}
    \renewcommand{\arraystretch}{1.16}
    \begin{tabularx}{\linewidth}{llX}
        \toprule
        \textbf{Organization} & \textbf{Readout} & \textbf{Evidence order and readout} \\
        \midrule
        \rowcolor{EviBlueTint} Semantic & Trailing & $E$\texttt{<ENT>} $A$\texttt{<ATT>} $R$\texttt{<REL>} $D$\texttt{<DET>} $S$\texttt{<SUM>}; \texttt{<SUM>} state only \\
        \rowcolor{EviBlueTint} Semantic & Distributed & $\mathcal U(E\,|\,A\,|\,R\,|\,D\,|\,S)$; five inserted-token states \\
        \rowcolor{EviBlueTint} Semantic & Boundary & $E$\texttt{<ENT>} $A$\texttt{<ATT>} $R$\texttt{<REL>} $D$\texttt{<DET>} $S$\texttt{<SUM>} \\
        \midrule
        \rowcolor{EviCoralTint} Mixed & Trailing & $D$\texttt{<ENT>} $S$\texttt{<ATT>} $E$\texttt{<REL>} $A$\texttt{<DET>} $R$\texttt{<SUM>}; \texttt{<SUM>} state only \\
        \rowcolor{EviCoralTint} Mixed & Distributed & $\mathcal U(D\,|\,S\,|\,E\,|\,A\,|\,R)$; five inserted-token states \\
        \rowcolor{EviCoralTint} Mixed & Boundary & $D$\texttt{<ENT>} $S$\texttt{<ATT>} $E$\texttt{<REL>} $A$\texttt{<DET>} $R$\texttt{<SUM>} \\
        \bottomrule
    \end{tabularx}
    \vspace{3pt}
    \parbox{\linewidth}{\small $E$: cookies, pastries, and display cases; $A$: warm indoor bakery; $R$: cookies on counter and pastries below; $D$: colorful icing and tiered glass displays; $S$: bakery counter with cookies. Each abbreviation includes its textual role label (e.g., $E$ includes \texttt{[Entity]}), which moves with the span in Mixed. $\mathcal U(X)$ inserts \texttt{<ENT>}, \texttt{<ATT>}, \texttt{<REL>}, \texttt{<DET>}, and \texttt{<SUM>} after the first $\lfloor T/5\rfloor,\ldots,\lfloor5T/5\rfloor$ tokens of $X$, respectively.}
\end{table}

\section{Additional Analyses}

\subsection{Boundary-Format Validity}
\label{app:boundary_format_validity}

We evaluate 1,000 query generations from each of the 12 MMEB retrieval tasks using the final Qwen3-VL-8B model trained on 500K pairs. As shown in Table~\ref{tab:boundary_format_validity_500k}, 11,942 of 12,000 generations (99.52\%) contain all five boundary tokens exactly once and in the prescribed order. The remaining 58 generations are incomplete; none contains duplicated or reordered boundary tokens.

\begin{table}[htbp]
    \centering
    \caption{Query-side boundary-format validity of the final 500K EviAlign model. A valid output contains each of the five boundary tokens exactly once and in the prescribed order.}
    \label{tab:boundary_format_validity_500k}
    \footnotesize
    \setlength{\tabcolsep}{7pt}
    \renewcommand{\arraystretch}{1.05}
    \begin{tabular}{lrrr}
        \toprule
        \textbf{Dataset} & \textbf{Queries} & \makecell{\textbf{Valid}\\\textbf{outputs}} & \makecell{\textbf{Valid format}\\\textbf{(\%)}} \\
        \midrule
        \multicolumn{4}{l}{\cellcolor{EviLavenderTint}\textit{In-domain}} \\
        VisDial & 1,000 & 998 & 99.8 \\
        CIRR & 1,000 & 994 & 99.4 \\
        VisualNews-t2i & 1,000 & 991 & 99.1 \\
        VisualNews-i2t & 1,000 & 997 & 99.7 \\
        MSCOCO-t2i & 1,000 & 993 & 99.3 \\
        MSCOCO-i2t & 1,000 & 999 & 99.9 \\
        NIGHTS & 1,000 & 995 & 99.5 \\
        WebQA & 1,000 & 992 & 99.2 \\
        \midrule
        \multicolumn{4}{l}{\cellcolor{EviLavenderTint}\textit{Out-of-domain}} \\
        FashionIQ & 1,000 & 998 & 99.8 \\
        Wiki-SS-NQ & 1,000 & 994 & 99.4 \\
        OVEN & 1,000 & 998 & 99.8 \\
        EDIS & 1,000 & 993 & 99.3 \\
        \midrule
        \rowcolor{EviBlueTint}\textbf{Overall} & \textbf{12,000} & \textbf{11,942} & \textbf{99.52} \\
        \bottomrule
    \end{tabular}
\end{table}

\subsection{Representation Analysis of Evidence-Aligned Readouts}
\label{app:leave_one_out}

Starting from the final 500K EviAlign model, we evaluate each evidence-aligned readout alone. We also remove one readout at a time and form the embedding by averaging the remaining four followed by L2 normalization, without retraining. Table~\ref{tab:readout_complementarity_500k} summarizes the 12-task averages: the individual states carry retrieval information, but none matches the full five-state aggregate. Removing any one readout lowers the overall average by 0.23--0.36 points. Tables~\ref{tab:ablation_single_slot} and~\ref{tab:leave_one_out_500k} give the corresponding per-task results, showing how the five contextualized states contribute to the final single-vector representation.

\begin{table}[htbp]
    \centering
    \caption{Readout contribution on MMEB (average Recall@1, \%).}
    \label{tab:readout_complementarity_500k}
    \footnotesize
    \setlength{\tabcolsep}{5pt}
    \renewcommand{\arraystretch}{1.12}
    \begin{tabularx}{\linewidth}{l*{5}{>{\centering\arraybackslash}X}}
        \toprule
        \textbf{Setting} & \textbf{ENT} & \textbf{ATT} & \textbf{REL} & \textbf{DET} & \textbf{SUM} \\
        \midrule
        \rowcolor{EviBlueTint} One-slot & 76.12 & 75.60 & 74.94 & 73.52 & 74.37 \\
        \rowcolor{EviCoralTint} Leave-one-out & 76.58 & 76.63 & 76.71 & 76.66 & 76.68 \\
        \midrule
        \multicolumn{6}{c}{\cellcolor{EviLavenderTint}\textbf{Full (all five readouts): 76.94}} \\
        \bottomrule
    \end{tabularx}
\end{table}

\begin{table}[htbp]
    \centering
    \caption{Per-task single-readout results on MMEB (Recall@1, \%; 500K model). Each readout is evaluated without retraining; Full averages all five.}
    \label{tab:ablation_single_slot}
    \scriptsize
    \setlength{\tabcolsep}{2.5pt}
    \renewcommand{\arraystretch}{1.08}
    \begin{tabular*}{\linewidth}{@{\extracolsep{\fill}}lccccc>{\columncolor{EviBlueTint}}c}
        \toprule
        \textbf{Task} & \textbf{ENT} & \textbf{ATT} & \textbf{REL} & \textbf{DET} & \textbf{SUM} & \textbf{Full} \\
        \midrule
        VisDial & 85.5 & \textbf{85.7} & 84.3 & 84.3 & 81.4 & 85.5 \\
        CIRR & 74.7 & 75.2 & 75.8 & \textbf{78.6} & 75.7 & 77.2 \\
        VisualNews-t2i & 76.0 & 72.7 & 77.3 & 77.9 & 76.1 & \textbf{78.9} \\
        VisualNews-i2t & 82.7 & \textbf{84.8} & 78.1 & 77.6 & 79.1 & 83.3 \\
        MSCOCO-t2i & 81.0 & 79.9 & 80.1 & 79.3 & 79.2 & \textbf{82.1} \\
        MSCOCO-i2t & 79.2 & 78.7 & 75.2 & 74.2 & 76.3 & \textbf{79.8} \\
        NIGHTS & \textbf{75.4} & 73.7 & 72.7 & 72.6 & 74.2 & 74.6 \\
        WebQA & 90.5 & 90.0 & 89.9 & 88.5 & 90.5 & \textbf{91.0} \\
        \midrule
        FashionIQ & 31.7 & 31.7 & 31.9 & 30.0 & 29.5 & \textbf{32.8} \\
        Wiki-SS-NQ & \textbf{75.3} & 73.6 & 72.3 & 67.2 & 71.6 & 73.6 \\
        OVEN & 69.3 & 70.1 & 70.9 & 61.8 & 68.6 & \textbf{72.2} \\
        EDIS & 92.1 & 91.1 & 90.8 & 90.2 & 90.2 & \textbf{92.3} \\
        \midrule
        \textbf{Average} & 76.1 & 75.6 & 74.9 & 73.5 & 74.4 & \textbf{76.9} \\
        \bottomrule
    \end{tabular*}
\end{table}

\begin{table}[htbp]
    \centering
    \caption{Per-task leave-one-out results on MMEB (Recall@1, \%; 500K model). Each column omits one readout without retraining; Full averages all five.}
    \label{tab:leave_one_out_500k}
    \scriptsize
    \setlength{\tabcolsep}{2.5pt}
    \renewcommand{\arraystretch}{1.08}
    \begin{tabular*}{\linewidth}{@{\extracolsep{\fill}}lccccc>{\columncolor{EviBlueTint}}c}
        \toprule
        \textbf{Task} & \textbf{w/o ENT} & \textbf{w/o ATT} & \textbf{w/o REL} & \textbf{w/o DET} & \textbf{w/o SUM} & \textbf{Full} \\
        \midrule
        VisDial & 85.1 & 85.1 & 85.4 & 85.4 & 85.4 & \textbf{85.5} \\
        CIRR & 77.1 & 77.1 & 76.9 & 76.2 & 76.9 & \textbf{77.2} \\
        VisualNews-t2i & 78.8 & 78.6 & 78.5 & 78.1 & 78.8 & \textbf{78.9} \\
        VisualNews-i2t & 82.7 & 82.2 & 83.2 & 83.1 & 82.9 & \textbf{83.3} \\
        MSCOCO-t2i & 81.9 & 82.0 & 82.0 & 82.0 & 81.9 & \textbf{82.1} \\
        MSCOCO-i2t & 79.2 & 79.5 & 79.7 & 79.5 & 79.7 & \textbf{79.8} \\
        NIGHTS & 74.1 & 74.5 & 74.5 & 74.5 & 74.1 & \textbf{74.6} \\
        WebQA & 90.9 & 90.9 & 90.9 & 90.9 & 90.6 & \textbf{91.0} \\
        \midrule
        FashionIQ & 32.5 & 32.7 & 32.6 & 32.6 & 32.4 & \textbf{32.8} \\
        Wiki-SS-NQ & 73.0 & 73.0 & 73.3 & 73.4 & 73.5 & \textbf{73.6} \\
        OVEN & 71.8 & 71.8 & 71.3 & 72.0 & 71.9 & \textbf{72.2} \\
        EDIS & 91.9 & 92.1 & 92.2 & 92.2 & 92.1 & \textbf{92.3} \\
        \midrule
        \textbf{Average} & 76.58 & 76.63 & 76.71 & 76.66 & 76.68 & \textbf{76.94} \\
        \bottomrule
    \end{tabular*}
\end{table}

\subsubsection{Readout-Level Representation Geometry}
\label{app:readout_geometry}

We examine the five readouts before their aggregation into a single retrieval embedding, using EviAlign and two matched five-readout controls from the factorial study. For this diagnostic, each readout is L2-normalized as $\tilde e_k(x)=e_k(x)/\|e_k(x)\|_2$. Positive-pair same-readout alignment is the mean cosine similarity between $\tilde e_k(q)$ and $\tilde e_k(c^+)$, averaged over the five readouts and matched query--candidate pairs. We quantify within-input dispersion with a Gaussian-potential statistic adapted from \citet{Wang2020AlignmentUniformity}, applied to distinct readout pairs $i<j$ within the same input:
\begin{equation}
\mathcal{D}_{\mathrm{readout}} = \log\,\mathbb{E}_{x,\,i<j}\!\left[\exp\!\left(-2\|\tilde e_i(x)-\tilde e_j(x)\|_2^2\right)\right].
\end{equation}

\begin{table}[htbp]
    \centering
    \caption{Readout geometry under matched five-readout controls on MMEB. Dispersion is diagnostic (more negative means greater separation); Recall@1 is the 12-task average.}
    \label{tab:readout_geometry_500k}
    \footnotesize
    \setlength{\tabcolsep}{4pt}
    \renewcommand{\arraystretch}{1.08}
    \begin{tabular*}{\linewidth}{@{\extracolsep{\fill}}lccc}
        \toprule
        \textbf{Setting} & \makecell{\textbf{Positive-pair}\\\textbf{alignment}} & \makecell{\textbf{Within-input}\\\textbf{dispersion}} & \makecell{\textbf{Recall@1}\\\textbf{(\%)}} \\
        \midrule
        \rowcolor{EviCoralTint} Mixed + boundary & 0.53 & $-3.18$ & 74.55 \\
        \rowcolor{EviCoralTint} Semantic + distributed & 0.56 & $-3.34$ & 74.85 \\
        \rowcolor{EviBlueTint} Semantic + boundary (EviAlign) & \textbf{0.63} & $-3.27$ & \textbf{76.94} \\
        \bottomrule
    \end{tabular*}
\end{table}

As shown in Table~\ref{tab:readout_geometry_500k}, positive-pair alignment increases from $0.53$ and $0.56$ in the two controls to $0.63$ with semantic evidence and boundary readout, following the same ordering as retrieval performance. All three settings retain comparable within-input dispersion. Together with the leave-one-out results in Table~\ref{tab:leave_one_out_500k}, these measurements characterize the evidence-aligned readouts that contribute to EviAlign's aggregated representation.

\subsection{Per-Task Backbone Comparison}
\label{app:backbone_per_task}

Table~\ref{tab:backbone_per_task_500k} reports the complete per-task results of the Qwen2-VL-7B and Qwen3-VL-8B variants summarized in the main comparison, together with their in-domain, out-of-domain, and overall averages.

\begin{table}[htbp]
    \centering
    \caption{Per-task EviAlign results with Qwen2-VL-7B and Qwen3-VL-8B under the 500K training budget.}
    \label{tab:backbone_per_task_500k}
    \footnotesize
    \setlength{\tabcolsep}{4pt}
    \renewcommand{\arraystretch}{1.06}
    \begin{minipage}[t]{0.56\linewidth}
        \vspace{0pt}
        \begin{tabularx}{\linewidth}{X>{\columncolor{EviBlueTint}}c>{\columncolor{EviCoralTint}}c}
            \toprule
            \multicolumn{3}{c}{\textbf{In-domain}} \\
            \midrule
            \textbf{Dataset} & \textbf{Qwen3-8B} & \textbf{Qwen2-7B} \\
            \midrule
            VisDial & 85.5 & 84.7 \\
            CIRR & 77.2 & 76.3 \\
            VisualNews-t2i & 78.9 & 78.0 \\
            VisualNews-i2t & 83.3 & 82.5 \\
            MSCOCO-t2i & 82.1 & 81.2 \\
            MSCOCO-i2t & 79.8 & 78.9 \\
            NIGHTS & 74.6 & 73.7 \\
            WebQA & 91.0 & 90.3 \\
            \midrule
            \textbf{Average} & \textbf{81.6} & \textbf{80.7} \\
            \bottomrule
        \end{tabularx}
    \end{minipage}\hfill
    \begin{minipage}[t]{0.41\linewidth}
        \vspace{0pt}
        \renewcommand{\arraystretch}{1.39}
        \begin{tabularx}{\linewidth}{X>{\columncolor{EviBlueTint}}c>{\columncolor{EviCoralTint}}c}
            \toprule
            \multicolumn{3}{c}{\textbf{Out-of-domain}} \\
            \midrule
            \textbf{Dataset} & \textbf{Qwen3-8B} & \textbf{Qwen2-7B} \\
            \midrule
            FashionIQ & 32.8 & 31.6 \\
            Wiki-SS-NQ & 73.6 & 70.4 \\
            OVEN & 72.2 & 69.0 \\
            EDIS & 92.3 & 89.4 \\
            \midrule
            \textbf{Average} & \textbf{67.7} & \textbf{65.1} \\
            \midrule
            \textbf{Overall (12)} & \textbf{76.9} & \textbf{75.5} \\
            \bottomrule
        \end{tabularx}
    \end{minipage}
\end{table}

\subsection{Structured vs. Free-Form Training Targets}
\label{app:output_discriminability}
To characterize the generation supervision used by the controlled models, we compare corresponding structured-evidence and free-form CoT targets from the training data. We sample 1,000 MMEB training inputs in a stratified manner and use GPT-4o (``gpt-4o-2024-11-20'') as the judge with temperature 0.0. Each target is scored independently from 1 to 5 for retrieval relevance, faithfulness, and specificity. For each input and dimension, the target with the higher score is preferred, and equal scores are counted as ties. We report score-derived win rates among non-tied comparisons together with mean scores over all inputs. The prompt template is provided in Appendix~\ref{app:prompt}.

As shown in Figure~\ref{fig:e2_judge}, EviAlign's structured training targets receive higher scores than the free-form CoT training targets across all three dimensions. Models trained with the two target formats nevertheless remain nearly tied under the same trailing readout (73.74 vs.\ 73.75). Higher target-quality scores therefore do not, by themselves, translate into a better trailing-readout retrieval representation; the boundary readout lets the semantic organization contribute to embedding construction.

\paragraph{Granularity of Structured Evaluation.}
Related work uses structured criteria at finer granularity: CuRe verifies category-aware atomic claims in video captions~\citep{gao2026claimlevelrubricrewardsvideo}, and Step-wise Rubrics as Rewards attributes rubric items to individual reasoning steps~\citep{xie2026step}. StepSTEM evaluates multimodal reasoning processes by aligning predicted steps with reference solutions, including interleaved image--text chains~\citep{jin2026unveiling}. Here, the three target-level dimensions characterize the supervision, while the controlled retrieval comparisons measure how that supervision contributes to the learned embedding.

\begin{figure}[t]
    \centering
    \includegraphics[width=\linewidth]{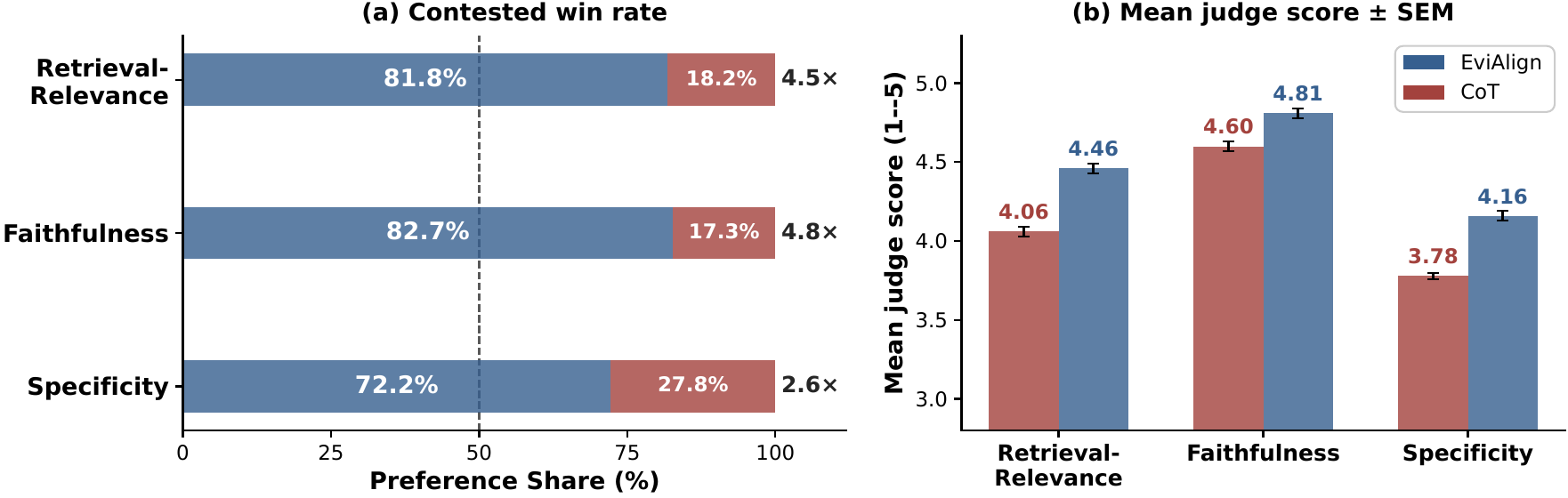}\caption{
    LLM-as-a-Judge comparison of EviAlign structured training targets and free-form CoT training targets.
    Win rates compare independently assigned 1--5 scores and exclude ties; mean scores use all sampled inputs.
    Multipliers denote the ratio of EviAlign wins to CoT wins among non-tied comparisons.
    }
    \label{fig:e2_judge}
\end{figure}

\subsection{Embedding Visualization}
\label{app:embedding_visualization}
\begin{figure}[t]
    \centering
    \includegraphics[width=0.92\linewidth]{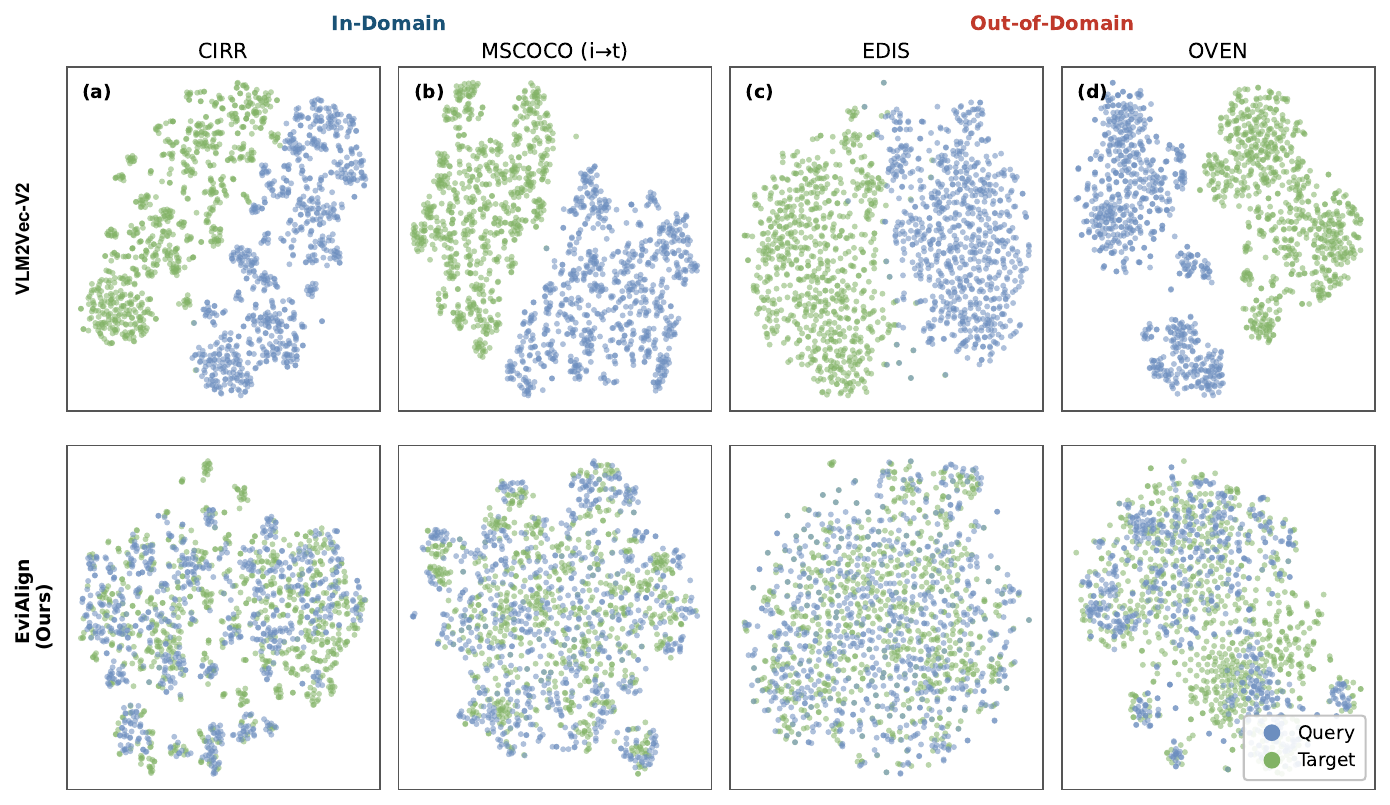}\caption{t-SNE visualization of query and target embeddings on four representative MMEB subsets.
    }
    \label{fig:tsne_qry_tgt}
\end{figure}

We further visualize query and target embeddings with t-SNE~\citep{Maaten2008TSNE} on four representative MMEB subsets: CIRR and MSCOCO image-to-text for in-domain evaluation, and EDIS and OVEN for out-of-domain evaluation. For each subset, 800 query embeddings and 800 target embeddings are jointly projected into a two-dimensional space. As shown in Figure~\ref{fig:tsne_qry_tgt}, VLM2Vec-V2 produces visibly separated query and target clusters, whereas EviAlign yields more interleaved query and target distributions across both in-domain and out-of-domain tasks.

\subsection{Attention Visualization}
\label{app:attention_visualization}
\begin{figure}[t]
    \centering
    \includegraphics[width=0.96\linewidth]{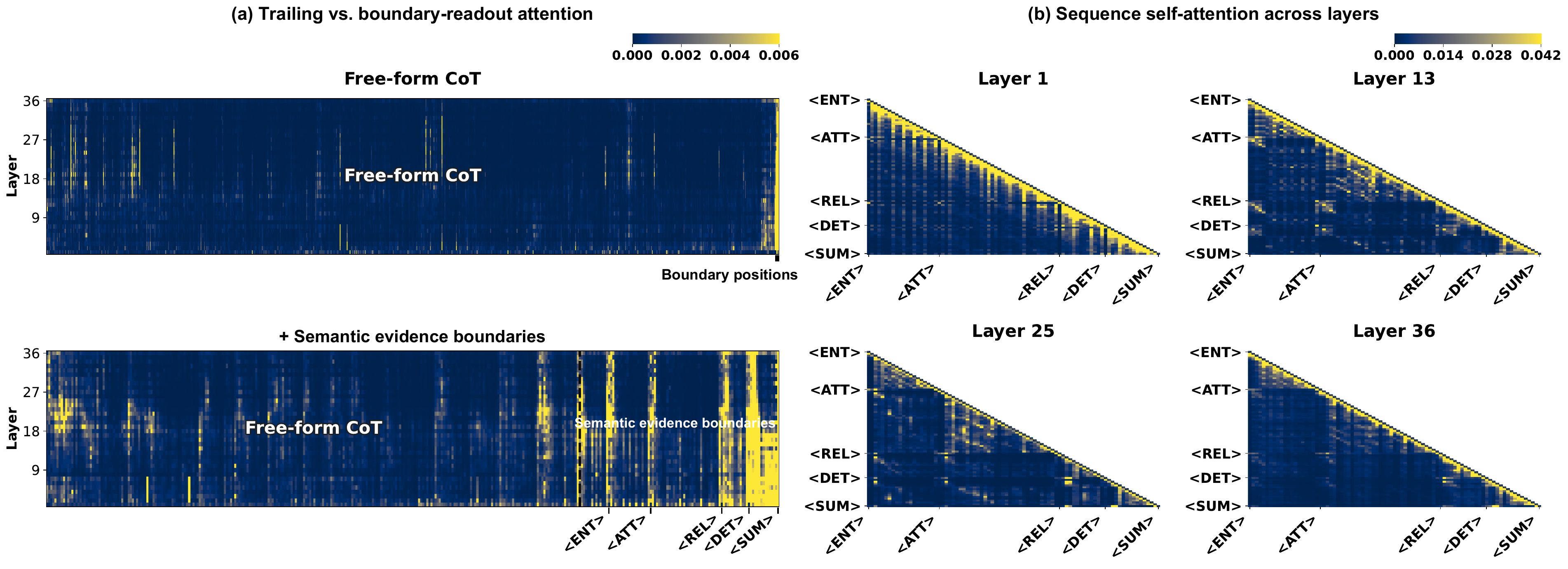}
    \caption{Attention patterns in the reasoning-prefix diagnostic. (a) Attention from the last evidence-boundary token (\texttt{<SUM>}) to preceding generated positions, compared with the trailing token of free-form CoT. (b) Layer-wise self-attention over sequence positions, with the five evidence-boundary positions marked. Panels (a) and (b) use separate color scales, shown above each panel.}
    \label{fig:attention_ablation}
\end{figure}

We visualize attention from the last evidence-boundary token (\texttt{<SUM>}) when free-form CoT precedes the semantic evidence, and compare it with the trailing readout of free-form CoT. Figure~\ref{fig:attention_ablation} shows attention concentrated near semantic evidence boundaries, with layer-wise patterns over the surrounding sequence positions. These qualitative patterns echo prior observations that specific token positions can serve as attention or information-flow anchors~\citep{Xiao2023EfficientSL,Barbero2025WhyDL,Wang2023LabelWA}.

\subsection{Data Scaling Analysis}
\label{app:data_scaling}
\begin{figure}[t]
    \centering
    \includegraphics[width=\linewidth]{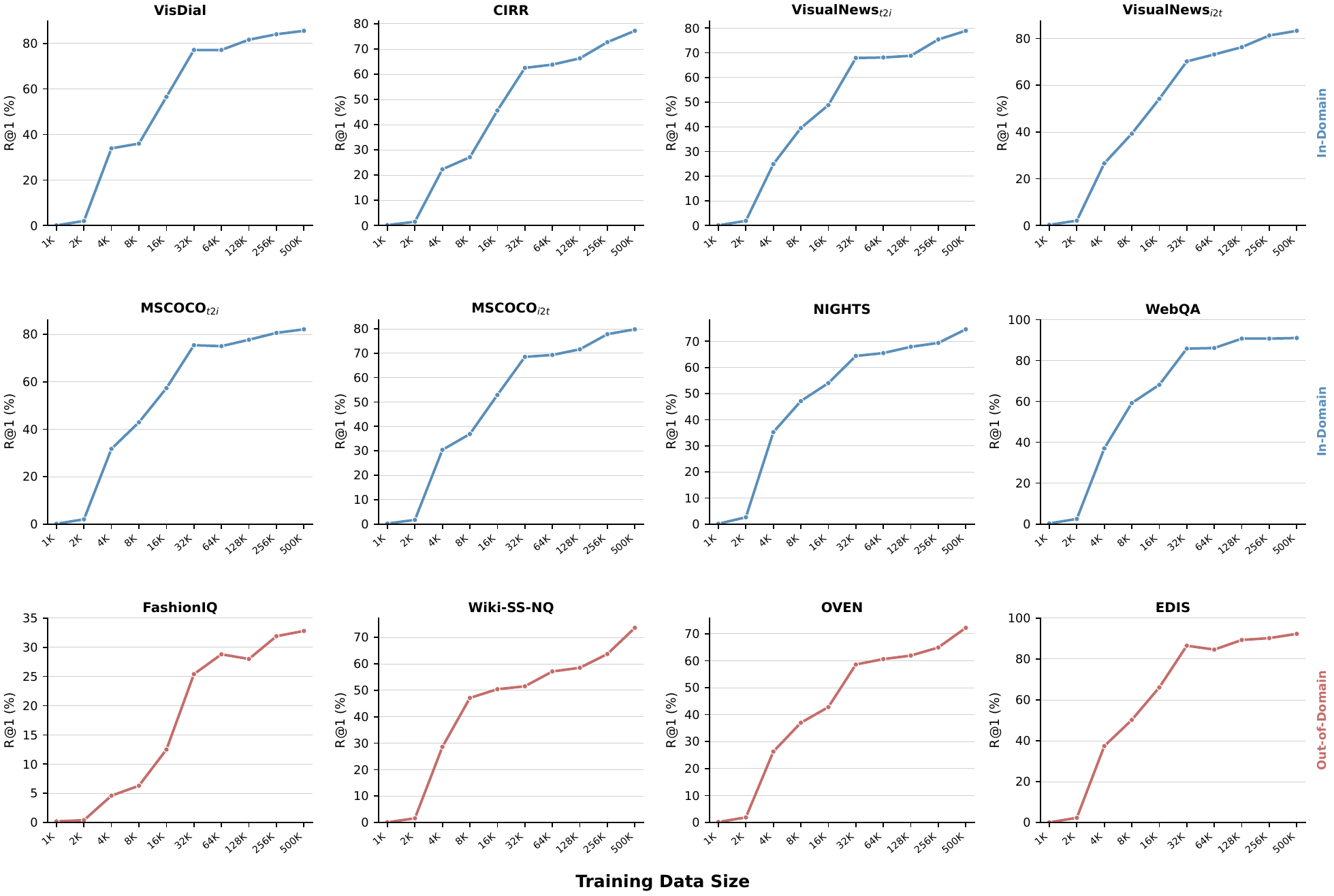}\caption{Data scaling curves of EviAlign on MMEB from 1K to 500K training pairs. We report Recall@1 across in-domain and out-of-domain retrieval benchmarks.
    }
    \label{fig:data_scaling}
\end{figure}

We further study how EviAlign scales with training data. Starting from the same training pool, we construct subsets ranging from 1K to 500K samples and train EviAlign with the same backbone, evidence configuration, and optimization settings. All models are evaluated on the same 12 MMEB retrieval tasks, including eight in-domain and four out-of-domain tasks.

As shown in Figure~\ref{fig:data_scaling}, performance remains low at 1K--2K training pairs. Inspection of generated outputs reveals frequent boundary-token omissions, causing the affected readouts to fall back to the final hidden state rather than the intended evidence boundaries. From 4K onward, the model progressively learns to follow the prescribed boundary-token format more reliably, allowing boundary-aligned readout to operate as intended. Alongside this transition, retrieval performance rises sharply and continues to improve with additional training data across both in-domain and out-of-domain tasks. Scaling from 256K to 500K further improves all 12 tasks, with particularly clear gains on CIRR, NIGHTS, Wiki-SS-NQ, and OVEN.

\subsection{Complexity and Scalability Analysis}
\label{app:complexity}
We analyze the two deployment stages at which EviAlign differs from conventional embedding systems: representation construction and index storage. The single-vector interface preserves index storage and similarity scoring; query-side evidence generation remains part of online encoding.

\subsubsection{Representation-Construction Efficiency}

EviAlign learns to generate semantic evidence under teacher forcing and reads the $K$ evidence-boundary states to construct its retrieval representation. At inference, it generates the evidence and aggregates these states into one embedding. Table~\ref{tab:efficiency_query_side} profiles this representation-construction path and a control that generates free-form CoT before the same semantic evidence while retaining the boundary readout. Under BF16 inference on NVIDIA H20 GPUs with a per-device batch size of 4, EviAlign reaches 76.94 average Recall@1 with 114.3 generated tokens on average, an end-to-end representation-construction time of 1,502.5 ms per batch, and 19.49 GB peak memory per GPU. Adding free-form CoT before the evidence yields 76.59 while increasing these costs to 702.1 tokens, 9,825.3 ms, and 44.46 GB. Both variants produce a single vector scored with cosine similarity.

\begin{table}[t]
\centering
\caption{Query-side representation-construction efficiency and retrieval quality. Recall@1 is averaged over the 12 MMEB retrieval tasks. Batch time measures end-to-end representation construction for four queries, including multimodal encoding, autoregressive generation when applicable, and embedding aggregation. Memory reports the peak GPU allocation under the same run.}
  \label{tab:efficiency_query_side}
  \small
  \setlength{\tabcolsep}{3.5pt}
  \renewcommand{\arraystretch}{1.15}
  \begin{adjustbox}{max width=\columnwidth}
  \begin{tabular}{lcccc}
      \toprule
      \textbf{Setting} & \textbf{Recall@1} & \textbf{Batch time} & \textbf{Tokens} & \textbf{Memory} \\
      \midrule
      VLM2Vec-V2 (2B) & 69.5 & 67.8 ms & -- & 17.9 GB \\
      GME-7B & 71.2 & 89.1 ms & -- & 15.5 GB \\
      Qwen3-VL-8B direct (ours) & 68.30 & 94.6 ms & -- & 17.4 GB \\
      \rowcolor{EviMintTint} EviAlign & \textbf{76.94} & 1,502.5 ms & 114.3 & 19.49 GB \\
      \rowcolor{EviCoralTint} Free-form CoT + semantic evidence & 76.59 & 9,825.3 ms & 702.1 & 44.46 GB \\
      \bottomrule
  \end{tabular}
  \end{adjustbox}
\end{table}

Generation length is also an optimization target in reasoning models: ExpThink uses experience-guided reinforcement learning for difficulty-adaptive chain-of-thought compression~\citep{bian2026expthink}. This targets the generation component of inference cost, complementing the distinction here between query-side representation construction and corpus-side index storage.

\subsubsection{Index Storage}
\label{app:deployment_storage}

EviAlign aggregates its five internal readouts into one dense vector before indexing. For a corpus of $|\mathcal{C}|$ items and embedding dimension $d$, it therefore stores $|\mathcal{C}|d$ scalars, without a multiplicative dependence on the number of evidence units. Table~\ref{tab:deployment_storage} calculates raw FP16 embedding storage for 1M items from the vector count and dimensionality of each configuration; compression, index metadata, and system-level overhead are excluded. EviAlign uses $d=4096$ with Qwen3-VL-8B and $d=2048$ with Qwen3-VL-2B. ColPali~\citep{Faysse2024ColPaliED} and ColBERT are included as reference multi-vector configurations. EviAlign remains compatible with standard dense-retrieval systems such as FAISS~\citep{johnson2019billion}.

\begin{table}[ht]
    \centering
    \caption{Calculated raw FP16 embedding storage for 1M retrieval items under the listed vector counts and dimensions; index overhead and compression are excluded.}
    \label{tab:deployment_storage}
    \small
    \setlength{\tabcolsep}{5pt}
    \renewcommand{\arraystretch}{1.2}
    \begin{adjustbox}{max width=\columnwidth}
    \begin{tabular}{l c c c c}
        \toprule
        \textbf{Method} & \textbf{Vecs/Doc} & \textbf{Dim} & \textbf{Storage/1M} & \textbf{Retrieval} \\
        \midrule
        ColPali & 1,030 & 128 & 263.7 GB & MaxSim \\
        ColBERT (text) & $\sim$30 & 128 & 7.7 GB & MaxSim \\
        \midrule
        EviAlign (8B) & 1 & 4,096 & 8.2 GB & Cosine \\
        EviAlign (2B) & 1 & 2,048 & 4.1 GB & Cosine \\
        \bottomrule
    \end{tabular}
    \end{adjustbox}
\end{table}

\subsection{Evidence Schema Details}
\label{app:slot_design}

EviAlign organizes generated evidence into explicit units with corresponding readout boundaries. Our implementation uses five units---Entity, Attribute, Relation, Detail, and Summary---as a compact empirical configuration. Entity covers objects or concepts together with identifying properties; Attribute captures scene-level characteristics; Relation describes actions, interactions, and spatial relations; Detail retains locally discriminative cues such as visible text, signs, textures, and patterns; and Summary provides a retrieval-focused global description. These roles follow the data-construction prompt used in our experiments.

These units provide stable semantic roles and explicit boundaries shared by generation and representation readout. Section~\ref{sec:representation_analysis} evaluates joint evidence-schema and readout configurations and supports the selected implementation among the tested settings.

\paragraph{Task-Specific Evidence Organization.}
The useful granularity of visual evidence depends on the decision it supports. In interactive settings, GameVerse evaluates how video-based reflection on failures and demonstrations informs subsequent gameplay~\citep{gameverse2026}; the generalist-game-player survey organizes the broader design space around datasets, models, harnesses, and benchmarks~\citep{zhang2026towards}. VSI-Super-Wild examines a different requirement: maintaining agent, object, and environment state across long, continuous real-world videos~\citep{VSI_Super_Wild}. EgoProx studies egocentric 3D proximity reasoning across intention, exploration, exploitation, and action sequences~\citep{Li_2026_CVPR}, while IPIBench evaluates how streaming assistants maintain and act on changing proactive requests~\citep{li2026ipibench}. Across these settings, relevant evidence depends on the spatial relation, evolving state, or user request being resolved. For retrieval, EviAlign organizes evidence around the intended match: identifying entities, scene attributes, relations, discriminative details, and a target-level summary. This task-conditioned organization provides the semantic units whose boundaries become the readout locations.

\paragraph{Evidence Use in Agentic Pipelines.}
The downstream consumer also shapes how intermediate information is used. Octopus dynamically orchestrates six capabilities for multimodal reasoning~\citep{guo2025octopus}, and ACE-Router uses interaction history to route among tools and agents~\citep{yao2026acerouter}. At the data-construction level, the ACE lens separates generated experience into environment, task, interaction, and verification components~\citep{zeng2026agenticdata}. These works distinguish the construction of intermediate information from the decisions it supports. EviAlign studies this distinction at the representation level: its evidence units supply explicit locations for reading the states that form a retrieval embedding.

\subsection{Case Studies}
\label{app:case_studies}
\newcommand{\evidenceunit}[2]{\begin{tcolorbox}[enhanced,breakable,sharp corners,
    colframe=blue!35!black,colback=blue!3!white,
    boxrule=0.5pt,arc=2pt,left=3pt,right=3pt,top=2pt,bottom=2pt,
    before skip=2pt,after skip=2pt]
  {\scriptsize\textbf{#1.} #2}
  \end{tcolorbox}}

Figure~\ref{fig:main_cirr_case} presents the CIRR composed-image example in the main text. Here we provide two additional cases from image-to-text retrieval on VisualNews and out-of-domain knowledge-grounded retrieval on OVEN. Each case presents the query and target first, followed by the generated evidence at readable single-column scale.

\begin{figure}[H]
\centering
\begin{tcolorbox}[enhanced,colframe=blue!35!black,colback=white,
  colbacktitle=blue!10!white,coltitle=black,fonttitle=\bfseries\small,
  title={Case 2 --- VisualNews: Image-to-Caption Retrieval},
  boxrule=0.8pt,arc=3pt,left=4pt,right=4pt,top=4pt,bottom=4pt]
\begin{minipage}[c]{0.48\linewidth}
\centering
\parbox[c][3.0cm][c]{\linewidth}{\centering\includegraphics[width=\linewidth,height=3.0cm,keepaspectratio]{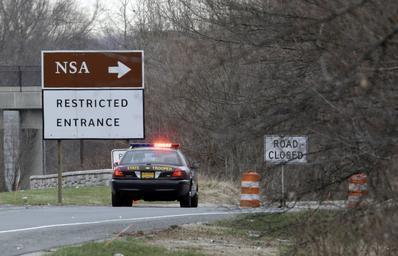}}\par
\vspace{2pt}{\scriptsize\textbf{Query image}}
\end{minipage}\hfill
\begin{minipage}[c]{0.48\linewidth}
\parbox[c][3.0cm][c]{\linewidth}{\raggedright\scriptsize\itshape A Maryland State Police cruiser sits at a blocked southbound entrance on the Baltimore--Washington Parkway that accesses Fort Meade.}\par
\vspace{2pt}{\scriptsize\textbf{Retrieved caption}}
\end{minipage}
\medskip
\evidenceunit{Entity}{brown rectangular NSA directional sign, white ``Restricted Entrance'' placard, Maryland State Police cruiser with emergency light bar, orange traffic cones}
\evidenceunit{Attribute}{outdoor, overcast daytime, rural road setting, law-enforcement / security checkpoint atmosphere}
\evidenceunit{Relation}{NSA sign and Restricted Entrance sign co-mounted on same roadside pole; police cruiser blocking road at entrance; traffic cones positioned along road edge}
\evidenceunit{Detail}{text: ``NSA'' with arrow on brown sign; text: ``Restricted Entrance'' on white sign; text: ``Road Closed'' on smaller placard; red/blue emergency light bar on police cruiser}
\evidenceunit{Summary}{Police cruiser blocking NSA restricted entrance on Baltimore--Washington Parkway near Fort Meade}
\end{tcolorbox}
\caption{VisualNews example: an image query, structured retrieval evidence, and the retrieved caption.}
\label{fig:case2}
\end{figure}

\begin{figure}[H]
\centering
\begin{tcolorbox}[enhanced,colframe=blue!35!black,colback=white,
  colbacktitle=blue!10!white,coltitle=black,fonttitle=\bfseries\small,
  title={Case 3 --- OVEN: Knowledge-Grounded Retrieval},
  boxrule=0.8pt,arc=3pt,left=4pt,right=4pt,top=4pt,bottom=4pt]
\begin{minipage}[c]{0.48\linewidth}
\begin{minipage}[c]{0.52\linewidth}
\centering
\parbox[c][3.0cm][c]{\linewidth}{\centering\includegraphics[width=\linewidth,height=3.0cm,keepaspectratio]{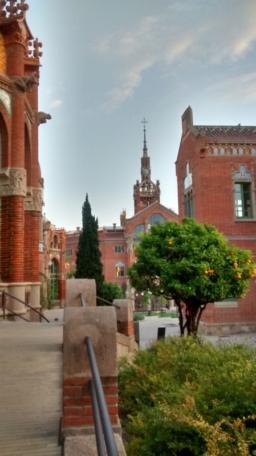}}\par
\vspace{2pt}{\scriptsize\textbf{Query image}}
\end{minipage}\hfill
\begin{minipage}[c]{0.44\linewidth}
\centering
\parbox[c][3.0cm][c]{\linewidth}{\makebox[\linewidth][c]{\scriptsize\itshape ``Where is this place?''}}\par
\vspace{2pt}{\scriptsize\textbf{Query text}}
\end{minipage}
\end{minipage}\hfill
\begin{minipage}[c]{0.48\linewidth}
\centering
\parbox[c][3.0cm][c]{\linewidth}{\centering\includegraphics[width=\linewidth,height=3.0cm,keepaspectratio]{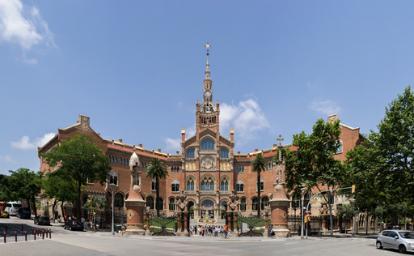}}\par
\vspace{2pt}{\scriptsize\textbf{Retrieved entry image}}
\end{minipage}
\medskip
\evidenceunit{Entity}{ornate red-brick pavilion buildings with Catalan Modernista sculptural facades, stone railings with brick bases, tall tree with yellow blossoms, paved garden walkway}
\evidenceunit{Attribute}{outdoor, daytime, clear sky, historic Catalan Modernista architectural style, well-maintained heritage garden setting}
\evidenceunit{Relation}{multi-pavilion complex enclosing a central garden; main tower with decorated dome visible at far end; stone railings lining the central walkway; paved path leading toward main building}
\evidenceunit{Detail}{polychrome tile and brick ornamental motifs on pavilion facades; conical spire with mosaic finish on central dome; decorative sculptural reliefs above archways; geometric stone-and-brick railing bases}
\evidenceunit{Summary}{Hospital de Sant Pau--style Catalan Modernista brick complex with ornate multi-pavilion layout and central heritage garden}
\end{tcolorbox}
\caption{OVEN example: the query combines an image and a place-identification question. The retrieved candidate is an image--text knowledge entry; its image is shown.}
\label{fig:case3}
\end{figure}

\paragraph{Case 2: Cross-Modal Image-to-Text Retrieval (VisualNews).}
Image-to-text retrieval connects visual evidence with a natural-language caption. Entity records the roadside signs, police cruiser, and traffic cones; Detail captures text such as ``NSA,'' ``Restricted Entrance,'' and ``Road Closed''; and Relation describes the blocked entrance. Summary integrates these cues into a description consistent with the retrieved caption.

\paragraph{Case 3: Knowledge-Grounded Retrieval (OVEN).}
This case uses an out-of-domain query format that asks the model to identify a place from an image. The retrieved candidate is a knowledge entry with both image and text; the figure shows its image. Entity and Attribute describe the architectural components and Catalan Modernista style, Relation preserves the multi-pavilion layout, and Detail records tile patterns, mosaic spires, and sculptural reliefs. Summary integrates these local and global cues into a landmark-level description of the retrieved entry.

\Needspace{8\baselineskip}
\section{Prompt Templates}
\label{app:prompt}
We document four categories of prompt templates used in experiments: (1)~the inference prompt of EviAlign, applied identically to every query and candidate, (2)~the data-construction prompt of EviAlign-Evidence, used to organize semantic-evidence targets, (3)~the prompts used in evidence-configuration ablations, and (4)~the judgment prompt used in the evidence-quality analysis.

\subsection{Inference Prompt of EviAlign}
\label{app:prompt_inference}

The evidence-generation template is shared across all 12 MMEB datasets. \texttt{\{\{Multimodal Input\}\}} contains the input text and/or image together with its retrieval task instruction. The ASSISTANT OUTPUT block shows the structure of the model-generated evidence. The five boundary tokens \texttt{<ENT>}, \texttt{<ATT>}, \texttt{<REL>}, \texttt{<DET>}, and \texttt{<SUM>} are registered as new vocabulary items initialized from the \texttt{[EOS]} embedding; their final-layer states provide the five boundary readouts (Section~\ref{sec:method_evidence_extract}).

\begingroup
\small
\begin{tcolorbox}[
    enhanced, breakable,
    title={\normalsize\textbf{Inference Prompt of EviAlign}},
    colframe=blue!45!black, colback=blue!2!white,
    colbacktitle=blue!10!white, coltitle=blue!50!black,
    fonttitle=\bfseries, halign title=center,
    boxrule=1pt, arc=4pt,
    left=6pt, right=6pt, top=5pt, bottom=5pt,
]
  \begin{tcolorbox}[
      enhanced, sharp corners,
      colframe=gray!50, colback=gray!8,
      title=\small\textbf{USER},
      coltitle=black, colbacktitle=gray!20, fonttitle=\small\bfseries,
      boxrule=0.5pt, left=5pt, right=5pt, top=3pt, bottom=3pt,
  ]
    \texttt{\{\{Multimodal Input\}\}}\\[4pt]
    Represent the above input texts, images, or any combination of the two as embeddings.
    Directly summarize the input in a structured format, using the tokens
    \texttt{<ENT>}, \texttt{<ATT>}, \texttt{<REL>}, \texttt{<DET>}, \texttt{<SUM>}
    to respectively represent the Entity, Attribute, Relation, Detail, and Summary.
  \end{tcolorbox}
  \smallskip
  \begin{tcolorbox}[
      enhanced, sharp corners,
      colframe=teal!55!black, colback=teal!4!white,
      title=\small\textbf{ASSISTANT OUTPUT},
      coltitle=black, colbacktitle=teal!15!white, fonttitle=\small\bfseries,
      boxrule=0.5pt, left=5pt, right=5pt, top=3pt, bottom=3pt,
  ]
    \texttt{[Entity] \{\ldots\}\;<ENT>}\\
    \texttt{[Attribute] \{\ldots\}\;<ATT>}\\
    \texttt{[Relation] \{\ldots\}\;<REL>}\\
    \texttt{[Detail] \{\ldots\}\;<DET>}\\
    \texttt{[Summary] \{\ldots\}\;<SUM>}
  \end{tcolorbox}
\end{tcolorbox}
\endgroup

\subsection{Training Data Construction Prompt of EviAlign-Evidence}
\label{app:prompt_construction}

We use GLM-4.1V-9B-Thinking~\citep{Hong2025GLM41VThinkingTV} to convert each training input from the MMEB-V1~\citep{Jiang2024VLM2VecTV} retrieval subset into five-unit semantic-evidence targets. The resulting annotations are filtered to remove excessive repetition, invalid formats, or outputs exceeding the 8192-token hard safety limit and are associated with approximately 500K query--candidate training pairs used for supervised fine-tuning.

\begingroup
\small
\begin{tcolorbox}[
    enhanced, breakable,
    title={\normalsize\textbf{Training Data Construction Prompt of EviAlign-Evidence}},
    colframe=blue!45!black, colback=blue!2!white,
    colbacktitle=blue!10!white, coltitle=blue!50!black,
    fonttitle=\bfseries, halign title=center,
    boxrule=1pt, arc=4pt,
    left=6pt, right=6pt, top=5pt, bottom=5pt,
]
  \begin{tcolorbox}[
      enhanced, sharp corners,
      colframe=gray!50, colback=gray!8,
      title=\small\textbf{USER},
      coltitle=black, colbacktitle=gray!20, fonttitle=\small\bfseries,
      boxrule=0.5pt, left=5pt, right=5pt, top=3pt, bottom=3pt,
  ]
    \texttt{\{\{Multimodal Input \}\}}\\[5pt]
    The above contains a retrieval input (text, image, or both).
    Use them for reference, but verify all details against the original input---the input may contain errors.
    Identify the retrieval task instruction (e.g., ``find a matching image'', ``find a similar image with the described changes'') and focus extraction on information most relevant to that task.\\[5pt]
    Extract key information and output \textbf{EXACTLY} in this format (5 lines only):\\[4pt]
    \texttt{[Entity]}\quad\textit{entities with bound identifying attributes; every entity needs at least one attribute (e.g., ``orange tabby cat'', not ``cat'')}\quad\texttt{<ENT>}\\[2pt]
    \texttt{[Attribute]}\quad\textit{scene-wide attributes only: lighting, camera angle, weather, time of day, image style, mood}\quad\texttt{<ATT>}\\[2pt]
    \texttt{[Relation]}\quad\textit{actions and spatial relationships (e.g., ``cat sitting on counter'', ``bus parked at hillside'')}\quad\texttt{<REL>}\\[2pt]
    \texttt{[Detail]}\quad\textit{discriminative visual cues: visible text as \texttt{text:``...''}, logos as \texttt{logo:``...''}; also patterns, materials, signs, etc.}\quad\texttt{<DET>}\\[2pt]
    \texttt{[Summary]}\quad\textit{concise summary of the retrieval target}\quad\texttt{<SUM>}\\[5pt]
    Rules:\;
    (i)~Output only the 5 lines---no commentary or analysis.\\[2pt]
    (ii)~\texttt{[Entity]} binds attributes: ``red dress'' in Entity, not ``dress'' + ``red'' in Attribute.\\[2pt]
    (iii)~\texttt{[Attribute]} is scene-wide only, never entity-specific.\\[2pt]
    (iv)~\texttt{[Detail]} must be grounded in the actual input; output \texttt{[]} only if truly absent.\\[2pt]
    (v)~\texttt{[Summary]} must be specific---avoid generic descriptions.
  \end{tcolorbox}
  \smallskip
  \begin{tcolorbox}[
      enhanced, sharp corners,
      colframe=teal!55!black, colback=teal!4!white,
      title=\small\textbf{ASSISTANT},
      coltitle=black, colbacktitle=teal!15!white, fonttitle=\small\bfseries,
      boxrule=0.5pt, left=5pt, right=5pt, top=3pt, bottom=3pt,
  ]
    \texttt{[Entity] \{\ldots\}\;<ENT>}\\
    \texttt{[Attribute] \{\ldots\}\;<ATT>}\\
    \texttt{[Relation] \{\ldots\}\;<REL>}\\
    \texttt{[Detail] \{\ldots\}\;<DET>}\\
    \texttt{[Summary] \{\ldots\}\;<SUM>}
  \end{tcolorbox}
\end{tcolorbox}
\endgroup

\subsection{Evidence-Configuration Ablation Prompts}
\label{app:prompt_slot7}

For the $K{=}7$ ablation, we retain the five default unit names and insert Context (\texttt{<CTX>}, scene context) and Emotion (\texttt{<EMO>}, emotional tone) between Detail and Summary. Mood is assigned to Emotion rather than Attribute in this template. The $K{=}1$ and $K{=}3$ inference prompts remove the unused units from the standard EviAlign template. Data construction and inference otherwise follow the same format.

\begingroup
  \small
  \begin{tcolorbox}[
      enhanced, breakable,
      title={\normalsize\textbf{Inference Prompt in Ablation Studies ($K{=}7$)}},
      colframe=blue!45!black, colback=blue!2!white,
      colbacktitle=blue!10!white, coltitle=blue!50!black,
      fonttitle=\bfseries, halign title=center,
      boxrule=1pt, arc=4pt,
      left=6pt, right=6pt, top=5pt, bottom=5pt,
  ]
    \begin{tcolorbox}[
        enhanced, sharp corners,
        colframe=gray!50, colback=gray!8,
        title=\small\textbf{USER},
        coltitle=black, colbacktitle=gray!20, fonttitle=\small\bfseries,
        boxrule=0.5pt, left=5pt, right=5pt, top=3pt, bottom=3pt,
    ]
      \texttt{\{\{Multimodal Input\}\}}\\[4pt]
      Represent the above input texts, images, or any combination of the two as embeddings.
      Directly summarize the input in a structured format, using the tokens
      \texttt{<ENT>}, \texttt{<ATT>}, \texttt{<REL>}, \texttt{<DET>}, \texttt{<CTX>}, \texttt{<EMO>}, \texttt{<SUM>}
      to respectively represent the Entity, Attribute, Relation, Detail, Context, Emotion, and Summary.
    \end{tcolorbox}
    \smallskip
    \begin{tcolorbox}[
        enhanced, sharp corners,
        colframe=teal!55!black, colback=teal!4!white,
        title=\small\textbf{ASSISTANT},
        coltitle=black, colbacktitle=teal!15!white, fonttitle=\small\bfseries,
        boxrule=0.5pt, left=5pt, right=5pt, top=3pt, bottom=3pt,
    ]
      \texttt{[Entity] \{\ldots\}\;<ENT>}\\
      \texttt{[Attribute] \{\ldots\}\;<ATT>}\\
      \texttt{[Relation] \{\ldots\}\;<REL>}\\
      \texttt{[Detail] \{\ldots\}\;<DET>}\\
      \texttt{[Context] \{\ldots\}\;<CTX>}\\
      \texttt{[Emotion] \{\ldots\}\;<EMO>}\\
      \texttt{[Summary] \{\ldots\}\;<SUM>}
    \end{tcolorbox}
  \end{tcolorbox}
\endgroup

\begingroup
  \small
  \begin{tcolorbox}[
      enhanced, breakable,
      title={\normalsize\textbf{Training Data Construction Prompt in Ablation Studies ($K{=}7$)}},
      colframe=blue!45!black, colback=blue!2!white,
      colbacktitle=blue!10!white, coltitle=blue!50!black,
      fonttitle=\bfseries, halign title=center,
      boxrule=1pt, arc=4pt,
      left=6pt, right=6pt, top=5pt, bottom=5pt,
  ]
    \begin{tcolorbox}[
        enhanced, sharp corners,
        colframe=gray!50, colback=gray!8,
        title=\small\textbf{USER},
        coltitle=black, colbacktitle=gray!20, fonttitle=\small\bfseries,
        boxrule=0.5pt, left=5pt, right=5pt, top=3pt, bottom=3pt,
    ]
      \texttt{\{\{Multimodal Input \}\}}\\[5pt]
      The above contains a retrieval input (text, image, or both).
      Use them for reference, but verify all details against the original input---the input may contain errors.
      Identify the retrieval task instruction (e.g., ``find a matching image'', ``find a similar image with the described changes'') and focus extraction on information most relevant to that task.\\[5pt]
      Extract key information and output \textbf{EXACTLY} in this format (7 lines only):\\[4pt]
      \texttt{[Entity]}\quad\textit{entities with bound identifying attributes; every entity needs at least one attribute (e.g., ``orange tabby cat'', not ``cat'')}\quad\texttt{<ENT>}\\[2pt]
      \texttt{[Attribute]}\quad\textit{scene-wide attributes only: lighting, camera angle, weather, time of day, image style}\quad\texttt{<ATT>}\\[2pt]
      \texttt{[Relation]}\quad\textit{actions and spatial relationships (e.g., ``cat sitting on counter'', ``bus parked at hillside'')}\quad\texttt{<REL>}\\[2pt]
      \texttt{[Detail]}\quad\textit{discriminative visual cues: visible text as \texttt{text:``...''}, logos as \texttt{logo:``...''}; also patterns, materials, signs, etc.}\quad\texttt{<DET>}\\[2pt]
      \texttt{[Context]}\quad\textit{situational or event background: where/when/what activity frames the scene (e.g., ``outdoor festival'', ``sports match at stadium'')---not entity-level traits}\quad\texttt{<CTX>}\\[2pt]
      \texttt{[Emotion]}\quad\textit{mood, sentiment, or subjective feeling conveyed by the scene (e.g., ``cheerful'', ``tense'', ``somber'')---must not duplicate neutral scene descriptions in Attribute}\quad\texttt{<EMO>}\\[2pt]
      \texttt{[Summary]}\quad\textit{concise summary of the retrieval target}\quad\texttt{<SUM>}\\[5pt]
      Rules:\;
      (i)~Output only the 7 lines---no commentary or analysis.\\[2pt]
      (ii)~\texttt{[Entity]} binds attributes: ``red dress'' in Entity, not ``dress'' + ``red'' in Attribute.\\[2pt]
      (iii)~\texttt{[Attribute]} is scene-wide only, never entity-specific.\\[2pt]
      (iv)~\texttt{[Context]} covers narrative situation, not physical attributes; output \texttt{[]} if the scene has no discernible event context.\\[2pt]
      (v)~\texttt{[Emotion]} covers evaluative or affective tone only; output \texttt{[]} if the scene is affectively neutral.\\[2pt]
      (vi)~\texttt{[Summary]} must be specific---avoid generic descriptions.
    \end{tcolorbox}
    \smallskip
    \begin{tcolorbox}[
        enhanced, sharp corners,
        colframe=teal!55!black, colback=teal!4!white,
        title=\small\textbf{ASSISTANT},
        coltitle=black, colbacktitle=teal!15!white, fonttitle=\small\bfseries,
        boxrule=0.5pt, left=5pt, right=5pt, top=3pt, bottom=3pt,
    ]
      \texttt{[Entity] \{\ldots\}\;<ENT>}\\
      \texttt{[Attribute] \{\ldots\}\;<ATT>}\\
      \texttt{[Relation] \{\ldots\}\;<REL>}\\
      \texttt{[Detail] \{\ldots\}\;<DET>}\\
      \texttt{[Context] \{\ldots\}\;<CTX>}\\
      \texttt{[Emotion] \{\ldots\}\;<EMO>}\\
      \texttt{[Summary] \{\ldots\}\;<SUM>}
    \end{tcolorbox}
\end{tcolorbox}
\endgroup

\Needspace{0.6\textheight}
\subsection{LLM-as-Judge Prompt}
\label{app:prompt_judge}

The judge independently evaluates each structured-evidence or free-form CoT training target for faithfulness, specificity, and retrieval relevance on a 1--5 scale. Per-input preferences are obtained by comparing the two scores for each dimension; equal scores are ties and are omitted from contested win rates.

\begingroup
  \small
  \begin{tcolorbox}[
      enhanced, breakable,
      title={\normalsize\textbf{LLM-as-Judge Prompt}},
      colframe=blue!45!black, colback=blue!2!white,
      colbacktitle=blue!10!white, coltitle=blue!50!black,
      fonttitle=\bfseries, halign title=center,
      boxrule=1pt, arc=4pt,
      left=6pt, right=6pt, top=5pt, bottom=5pt,
  ]
    \begin{tcolorbox}[
        enhanced, sharp corners,
        colframe=gray!50, colback=gray!8,
        title=\small\textbf{USER},
        coltitle=black, colbacktitle=gray!20, fonttitle=\small\bfseries,
        boxrule=0.5pt, left=5pt, right=5pt, top=3pt, bottom=3pt,
    ]
      You are evaluating the quality of a training target for a multimodal retrieval model.
      The target provides intermediate textual supervision before the retrieval embedding is constructed.\\[4pt]
      Rate the following training target on three dimensions (1--5 scale):\\[3pt]
      \textbf{1.~Faithfulness} (1--5): Does the target faithfully satisfy the retrieval requirements expressed by the input, including its text, images, and task instruction? Higher = more faithful to the input's retrieval requirements.\\[2pt]
      \textbf{2.~Specificity} (1--5): Does the target contain specific, discriminative information that would help distinguish the correct match from similar candidates? Higher = more specific and discriminative.\\[2pt]
      \textbf{3.~Retrieval-Relevance} (1--5): Is the target directly useful for matching this input to its correct target in a retrieval task? Higher = more retrieval-aligned; lower = contains verbose or irrelevant content.\\[4pt]
      \texttt{\{multimodal input\}}\\
      Training target to evaluate:\\
      \texttt{---}\\
      \texttt{\{trace\}}\\
      \texttt{---}\\[4pt]
      Output \textbf{ONLY} a JSON object:\\
      \texttt{\{"faithfulness": <int>, "specificity": <int>, "retrieval\_relevance": <int>\}}
    \end{tcolorbox}

\end{tcolorbox}
\endgroup


\begin{thebibliography}{59}
\providecommand{\natexlab}[1]{#1}
\providecommand{\url}[1]{\texttt{#1}}
\expandafter\ifx\csname urlstyle\endcsname\relax
  \providecommand{\doi}[1]{doi: #1}\else
  \providecommand{\doi}{doi: \begingroup \urlstyle{rm}\Url}\fi

\bibitem[Bai et~al.(2025)Bai, Cai, Chen, Chen, Chen, Cheng, Deng, Ding, Gao,
  Ge, Ge, Guo, Huang, Huang, Huang, Hui, Jiang, Li, Li, Li, Li, Lin, Lin, Liu,
  Liu, Liu, Liu, Liu, Liu, Lu, Luo, Lv, Men, Meng, Ren, Ren, Song, Sun, Tang,
  Tu, Wan, Wang, Wang, Wang, Wang, Xie, Xu, Xu, Xu, Yang, Yang, Yang, Yang, Yu,
  Zhang, Zhang, Zhang, Zheng, Zhong, Zhou, Zhou, Zhou, Zhu, and
  Zhu]{Bai2025Qwen3VLTR}
Shuai Bai, Yuxuan Cai, Ruizhe Chen, Keqin Chen, Xionghui Chen, Zesen Cheng,
  Lianghao Deng, Wei Ding, Chang Gao, Chunjiang Ge, Wenbin Ge, Zhifang Guo,
  Qidong Huang, Jie Huang, Fei Huang, Binyuan Hui, Shutong Jiang, Zhaohai Li,
  Mingsheng Li, Mei Li, Kaixin Li, Zicheng Lin, Junyang Lin, Xuejing Liu,
  Jiawei Liu, Chenglong Liu, Yang Liu, Dayiheng Liu, Shixuan Liu, Dunjie Lu,
  Ruilin Luo, Chenxu Lv, Rui Men, Lingchen Meng, Xuancheng Ren, Xingzhang Ren,
  Sibo Song, Yuchong Sun, Jun Tang, Jianhong Tu, Jianqiang Wan, Peng Wang,
  Pengfei Wang, Qiuyue Wang, Yuxuan Wang, Tianbao Xie, Yiheng Xu, Haiyang Xu,
  Jin Xu, Zhibo Yang, Mingkun Yang, Jianxin Yang, An~Yang, Bowen Yu, Fei Zhang,
  Hang Zhang, Xi~Zhang, Bo~Zheng, Humen Zhong, Jingren Zhou, Fan Zhou, Jing
  Zhou, Yuanzhi Zhu, and Ke~Zhu.
\newblock Qwen3-vl technical report.
\newblock \emph{arXiv preprint arXiv:2511.21631}, 2025.

\bibitem[Barbero et~al.(2025)Barbero, Arroyo, Gu, Perivolaropoulos, Bronstein,
  Veli{\v{c}}kovi{\'c}, and Pascanu]{Barbero2025WhyDL}
Federico Barbero, {\'A}lvaro Arroyo, Xiangming Gu, Christos Perivolaropoulos,
  Michael Bronstein, Petar Veli{\v{c}}kovi{\'c}, and Razvan Pascanu.
\newblock Why do llms attend to the first token?
\newblock \emph{arXiv preprint arXiv:2504.02732}, 2025.

\bibitem[Bian et~al.(2026)Bian, Zhang, Jin, Luo, Cheng, Wang, Jiang, and
  Wang]{bian2026expthink}
Tingcheng Bian, Yuzhe Zhang, Jing Jin, Jinchang Luo, MingQuan Cheng, Haiwei
  Wang, Wenyuan Jiang, and Miaohui Wang.
\newblock {ExpThink}: Experience-guided reinforcement learning for adaptive
  chain-of-thought compression.
\newblock \emph{arXiv preprint arXiv:2605.07501}, 2026.
\newblock URL \url{https://arxiv.org/abs/2605.07501}.

\bibitem[Chang et~al.(2022)Chang, Narang, Suzuki, Cao, Gao, and
  Bisk]{chang2022webqa}
Yingshan Chang, Mridu Narang, Hisami Suzuki, Guihong Cao, Jianfeng Gao, and
  Yonatan Bisk.
\newblock Webqa: Multihop and multimodal qa.
\newblock In \emph{Proceedings of the IEEE/CVF Conference on Computer Vision
  and Pattern Recognition}, pp.\  16495--16504, 2022.

\bibitem[Chen et~al.(2026)Chen, Ye, Zhang, Du, Pu, Wang, Zuo, Duan, Ma, and
  Zhang]{chen2026snapbench}
Zirong Chen, Fuda Ye, Kuan Zhang, Enjun Du, Junfu Pu, Xinlei Wang, Xinyu Zuo,
  Lisheng Duan, Jin Ma, and Yongqi Zhang.
\newblock {SnapBench}: Benchmarking snap-and-ask multimodal retrieval for
  mobile interactions.
\newblock \emph{arXiv preprint arXiv:2608.29607}, 2026.
\newblock URL \url{https://arxiv.org/abs/2608.29607}.

\bibitem[Cui et~al.(2026)Cui, Cheng, Chen, Shukla, Awasthi, Pan, Ahuja, Mishra,
  Tian, Guo, Lim, Singh, and Fan]{Cui2025ThinkTE}
Xuanming Cui, Jianpeng Cheng, Hong-You Chen, Satya~Narayan Shukla, Abhijeet
  Awasthi, Xichen Pan, Chaitanya Ahuja, Shlok~Kumar Mishra, Taipeng Tian,
  Qi~Guo, Ser-Nam Lim, Aashu Singh, and Xiangjun Fan.
\newblock Think then embed: Generative context improves multimodal embedding.
\newblock In \emph{The Fourteenth International Conference on Learning
  Representations}, 2026.

\bibitem[Das et~al.(2017)Das, Kottur, Gupta, Singh, Yadav, Moura, Parikh, and
  Batra]{das2017visual}
Abhishek Das, Satwik Kottur, Khushi Gupta, Avi Singh, Deshraj Yadav, Jos{\'e}
  M.~F. Moura, Devi Parikh, and Dhruv Batra.
\newblock Visual dialog.
\newblock In \emph{Proceedings of the IEEE Conference on Computer Vision and
  Pattern Recognition}, pp.\  326--335, 2017.

\bibitem[Du et~al.(2026{\natexlab{a}})Du, Liu, Chen, Zuo, Luo, Tao, Duan,
  Liang, Ma, Pu, and Zhang]{du2026evirank}
Enjun Du, Siyi Liu, Zirong Chen, Xinyu Zuo, Jinwen Luo, Ruiwen Tao, Lisheng
  Duan, Haijin Liang, Jin Ma, Junfu Pu, and Yongqi Zhang.
\newblock {EviRank}: Structured relevance evidence for multimodal image
  re-ranking.
\newblock In \emph{Proceedings of the 2026 Conference on Empirical Methods in
  Natural Language Processing}, 2026{\natexlab{a}}.
\newblock URL \url{https://arxiv.org/abs/2608.20886}.

\bibitem[Du et~al.(2026{\natexlab{b}})Du, Liu, Zheng, Li, Guo, Zhang, and
  Zou]{du2026omni}
Enjun Du, Siyi Liu, Ziyu Zheng, Jingyu Li, Yiwen Guo, Yongqi Zhang, and Difan
  Zou.
\newblock {Omni-Streaming Thinking}.
\newblock \emph{arXiv preprint arXiv:2609.15128}, 2026{\natexlab{b}}.
\newblock URL \url{https://arxiv.org/abs/2609.15128}.

\bibitem[Du et~al.(2026{\natexlab{c}})Du, Zhou, Du, Liu, Chen, Zheng, and
  Zhang]{du2026ledgermind}
Enjun Du, Hange Zhou, Chenxu Du, Siyi Liu, Zirong Chen, Ziyu Zheng, and Yongqi
  Zhang.
\newblock {LEDGERMIND}: Provenance-constrained multimodal agentic reasoning
  with a structured evidence ledger.
\newblock \emph{arXiv preprint arXiv:2607.28374}, 2026{\natexlab{c}}.
\newblock URL \url{https://arxiv.org/abs/2607.28374}.

\bibitem[Faysse et~al.(2025)Faysse, Sibille, Wu, Omrani, Viaud, Hudelot, and
  Colombo]{Faysse2024ColPaliED}
Manuel Faysse, Hugues Sibille, Tony Wu, Bilel Omrani, Gautier Viaud, C{\'e}line
  Hudelot, and Pierre Colombo.
\newblock Colpali: Efficient document retrieval with vision language models.
\newblock In \emph{The Thirteenth International Conference on Learning
  Representations}, 2025.

\bibitem[Fu et~al.(2023)Fu, Tamir, Sundaram, Chai, Zhang, Dekel, and
  Isola]{fu2023dreamsim}
Stephanie Fu, Netanel Tamir, Shobhita Sundaram, Lucy Chai, Richard Zhang, Tali
  Dekel, and Phillip Isola.
\newblock Dreamsim: Learning new dimensions of human visual similarity using
  synthetic data.
\newblock In \emph{Advances in Neural Information Processing Systems},
  volume~36, pp.\  50742--50768, 2023.

\bibitem[Gao et~al.(2026)Gao, Dong, Chen, Zhong, Ruan, Hou, Chen, Hu, and
  Tang]{gao2026claimlevelrubricrewardsvideo}
Mingqi Gao, Hongyuan Dong, Yifei Chen, Zhisheng Zhong, Zheng Ruan, Wenjin Hou,
  Yu~Chen, Han Hu, and Yansong Tang.
\newblock Claim-level rubric rewards for video caption reinforcement learning.
\newblock \emph{arXiv preprint arXiv:2607.05150}, 2026.
\newblock URL \url{https://arxiv.org/abs/2607.05150}.

\bibitem[{GLM-V Team} et~al.(2025){GLM-V Team}, Wang, Gan, Tang, Cheng, Qi, Ji,
  Pan, Duan, Wang, Wang, Cheng, He, Su, Yang, Pan, Zeng, Wang, Chen, Shi, Pang,
  Zhang, Yin, Yang, Chen, Li, Zhu, Chen, Xu, Xu, Chen, Lin, Chen, Wang, Chen,
  Lei, Gong, Pan, Liu, Xu, Zhang, Zheng, Lyu, Tu, Yang, Meng, Zhong, Huang,
  Zhao, Xue, Zhang, Luo, Hao, Tong, Jia, Li, Liu, Zhang, Lyu, Zhang, Fan,
  Huang, Xue, Wang, Wang, Wang, An, Du, Huang, Niu, Shi, Wang, Wang, Yue, Li,
  Liu, Zhang, Wang, Zhang, Xue, Du, Hou, Wang, Hong, Yu, Gu, Zhang, Liu, Xu,
  Li, Huang, Dong, and Tang]{Hong2025GLM41VThinkingTV}
{GLM-V Team}, Guo Wang, Guobing Gan, Haomiao Tang, Jiale Cheng, Ji~Qi, Junhui
  Ji, Lihang Pan, Shuaiqi Duan, Weihan Wang, Yan Wang, Yean Cheng, Zehai He,
  Zhe Su, Zhen Yang, Ziyang Pan, Aohan Zeng, Baoxu Wang, Bin Chen, Boyan Shi,
  Changyu Pang, Chenhui Zhang, Da~Yin, Fan Yang, Guoqing Chen, Haochen Li,
  Jiale Zhu, Jiali Chen, Jiaxing Xu, Jiazheng Xu, Jing Chen, Jinghao Lin,
  Jinhao Chen, Jinjiang Wang, Junjie Chen, Leqi Lei, Letian Gong, Leyi Pan,
  Mingdao Liu, Mingde Xu, Mingzhi Zhang, Qinkai Zheng, Ruiliang Lyu, Shangqin
  Tu, Sheng Yang, Shengbiao Meng, Shi Zhong, Shiyu Huang, Shuyuan Zhao, Siyan
  Xue, Tianshu Zhang, Tianwei Luo, Tianxiang Hao, Tianyu Tong, Wei Jia, Wenkai
  Li, Xiao Liu, Xiaohan Zhang, Xin Lyu, Xinyu Zhang, Xinyue Fan, Xuancheng
  Huang, Yadong Xue, Yanfeng Wang, Yanling Wang, Yanzi Wang, Yifan An, Yifan
  Du, Yiheng Huang, Yilin Niu, Yiming Shi, Yu~Wang, Yuan Wang, Yuanchang Yue,
  Yuchen Li, Yusen Liu, Yutao Zhang, Yuting Wang, Yuxuan Zhang, Zhao Xue,
  Zhengxiao Du, Zhenyu Hou, Zihan Wang, Wenyi Hong, Wenmeng Yu, Xiaotao Gu,
  Peng Zhang, Debing Liu, Bin Xu, Juanzi Li, Minlie Huang, Yuxiao Dong, and Jie
  Tang.
\newblock {GLM}-4.5{V} and {GLM}-4.1{V}-thinking: Towards versatile multimodal
  reasoning with scalable reinforcement learning.
\newblock \emph{arXiv preprint arXiv:2507.01006}, 2025.

\bibitem[Gu et~al.(2026{\natexlab{a}})Gu, Yang, Zhang, An, Feng, Zhang, Cai,
  Deng, and Bing]{Gu2025UniMEV2MF}
Tiancheng Gu, Kaicheng Yang, Kaichen Zhang, Xiang An, Ziyong Feng, Yueyi Zhang,
  Weidong Cai, Jiankang Deng, and Lidong Bing.
\newblock Unime-v2: Mllm-as-a-judge for universal multimodal embedding
  learning.
\newblock In \emph{Proceedings of the AAAI Conference on Artificial
  Intelligence}, volume~40, pp.\  21378--21386, 2026{\natexlab{a}}.

\bibitem[Gu et~al.(2026{\natexlab{b}})Gu, Xin, Zhang, Yang, Chua, Li, Zhang,
  Chen, Zhao, Xie, Liu, Lu, Han, Pavone, and Li]{VSI_Super_Wild}
Tianjun Gu, Tianyu Xin, Kuan Zhang, Bowen Yang, Kok-Chung Chua, Peize Li,
  Xinran Zhang, Yupeng Chen, Qiyue Zhao, Qinlei Xie, Jianhang Liu, Yucheng Lu,
  Yinan Han, Marco Pavone, and Yiming Li.
\newblock Towards spatial supersensing in the wild.
\newblock In \emph{The Nineteenth European Conference on Computer Vision},
  2026{\natexlab{b}}.
\newblock URL \url{https://arxiv.org/abs/2607.13681}.

\bibitem[G\"unther et~al.(2024)G\"unther, Mohr, Williams, Wang, and
  Xiao]{Gunther2024LateChunking}
Michael G\"unther, Isabelle Mohr, Daniel~James Williams, Bo~Wang, and Han Xiao.
\newblock Late chunking: Contextual chunk embeddings using long-context
  embedding models.
\newblock \emph{arXiv preprint arXiv:2409.04701}, 2024.

\bibitem[Guo et~al.(2025)Guo, Xu, Yao, Lu, Lin, Hu, Tang, Wang, and
  Chen]{guo2025octopus}
Yifu Guo, Zishan Xu, Zhiyuan Yao, Yuquan Lu, Jiaye Lin, Sen Hu, Zhenheng Tang,
  Huacan Wang, and Ronghao Chen.
\newblock {Octopus}: Agentic multimodal reasoning with six-capability
  orchestration.
\newblock \emph{arXiv preprint arXiv:2511.15351}, 2025.
\newblock URL \url{https://arxiv.org/abs/2511.15351}.

\bibitem[He et~al.(2026)He, Hao, Yang, Ma, Jia, Wu, Zhao, Guo, and
  Wang]{He2026PLUME}
Chenwei He, Xiangzhao Hao, Tianyu Yang, Yuxiang Ma, Yuheng Jia, Lingxiang Wu,
  Chaoyang Zhao, Haiyun Guo, and Jinqiao Wang.
\newblock Plume: Latent reasoning based universal multimodal embedding.
\newblock \emph{arXiv preprint arXiv:2604.02073}, 2026.

\bibitem[Hu et~al.(2023)Hu, Luan, Chen, Khandelwal, Joshi, Lee, Toutanova, and
  Chang]{hu2023open}
Hexiang Hu, Yi~Luan, Yang Chen, Urvashi Khandelwal, Mandar Joshi, Kenton Lee,
  Kristina Toutanova, and Ming-Wei Chang.
\newblock Open-domain visual entity recognition: Towards recognizing millions
  of wikipedia entities.
\newblock In \emph{Proceedings of the IEEE/CVF International Conference on
  Computer Vision}, pp.\  12065--12075, 2023.

\bibitem[Jiang et~al.(2025)Jiang, Meng, Yang, Yavuz, Zhou, and
  Chen]{Jiang2024VLM2VecTV}
Ziyan Jiang, Rui Meng, Xinyi Yang, Semih Yavuz, Yingbo Zhou, and Wenhu Chen.
\newblock {VLM}2vec: Training vision-language models for massive multimodal
  embedding tasks.
\newblock In \emph{The Thirteenth International Conference on Learning
  Representations}, 2025.

\bibitem[Jin et~al.(2025)Jin, Liu, Gao, Shi, Liang, Zheng, Kuang, Zeng, and
  Kan]{jin2025dynamic}
Jing Jin, Xu~Liu, Te~Gao, Zhihong Shi, Yixiong Liang, Ruiqing Zheng, Hulin
  Kuang, Min Zeng, and Shichao Kan.
\newblock Dynamic residual encoding with slide-level contrastive learning for
  end-to-end whole slide image representation.
\newblock In \emph{Proceedings of the 33rd ACM International Conference on
  Multimedia}, pp.\  8389--8398, 2025.
\newblock \doi{10.1145/3746027.3755469}.
\newblock URL \url{https://arxiv.org/abs/2511.05034}.

\bibitem[Jin et~al.(2026)Jin, Liu, Bai, Lou, Wang, Yuan, Chen, Zhu, Zeng, Zhu,
  Feng, and Xu]{jin2026unveiling}
Jing Jin, Hao Liu, Yan Bai, Yihang Lou, Zhenke Wang, Tianrun Yuan, Juntong
  Chen, Yongkang Zhu, Fanhu Zeng, Xuanyu Zhu, Tao Feng, and Yige Xu.
\newblock Unveiling fine-grained visual traces: Evaluating multimodal
  interleaved reasoning chains in multimodal {STEM} tasks.
\newblock \emph{arXiv preprint arXiv:2604.19697}, 2026.
\newblock URL \url{https://arxiv.org/abs/2604.19697}.

\bibitem[Johnson et~al.(2021)Johnson, Douze, and J{\'e}gou]{johnson2019billion}
Jeff Johnson, Matthijs Douze, and Herv{\'e} J{\'e}gou.
\newblock {Billion-Scale Similarity Search with {GPUs}}.
\newblock \emph{IEEE Transactions on Big Data}, 7\penalty0 (3):\penalty0
  535--547, 2021.

\bibitem[Khattab \& Zaharia(2020)Khattab and Zaharia]{Khattab2020ColBERTEA}
Omar Khattab and Matei Zaharia.
\newblock Colbert: Efficient and effective passage search via contextualized
  late interaction over bert.
\newblock In \emph{Proceedings of the 43rd International ACM SIGIR Conference
  on Research and Development in Information Retrieval}, pp.\  39--48, 2020.
\newblock \doi{10.1145/3397271.3401075}.

\bibitem[Kwiatkowski et~al.(2019)Kwiatkowski, Palomaki, Redfield, Collins,
  Parikh, Alberti, Epstein, Polosukhin, Devlin, Lee, Toutanova, Jones, Kelcey,
  Chang, Dai, Uszkoreit, Le, and Petrov]{Kwiatkowski2019NaturalQA}
Tom Kwiatkowski, Jennimaria Palomaki, Olivia Redfield, Michael Collins,
  Ankur~P. Parikh, Chris Alberti, Danielle Epstein, Illia Polosukhin, Jacob
  Devlin, Kenton Lee, Kristina Toutanova, Llion Jones, Matthew Kelcey, Ming-Wei
  Chang, Andrew~M. Dai, Jakob Uszkoreit, Quoc~V. Le, and Slav Petrov.
\newblock Natural questions: A benchmark for question answering research.
\newblock \emph{Transactions of the Association for Computational Linguistics},
  7:\penalty0 453--466, 2019.

\bibitem[Lan et~al.(2026)Lan, Niu, Meng, Zhou, and Su]{Lan2025UMER1ER}
Zhibin Lan, Liqiang Niu, Fandong Meng, Jie Zhou, and Jinsong Su.
\newblock Ume-r1: Exploring reasoning-driven generative multimodal embeddings.
\newblock In \emph{The Fourteenth International Conference on Learning
  Representations}, 2026.

\bibitem[Li et~al.(2026{\natexlab{a}})Li, Chen, Piao, Pan, Yu, Wang, Yan, Yue,
  Wang, Chen, Huang, and Liu]{Li_2026_CVPR}
Jinzhao Li, Yinuo Chen, Dongxu Piao, Panwang Pan, Yifan Yu, Dong Wang, Honglei
  Yan, Liang Yue, Shaofei Wang, Yixin Chen, Siyuan Huang, and Miao Liu.
\newblock {EgoProx}: Evaluating {MLLMs} on egocentric {3D} proximity reasoning
  across a cognitive hierarchy.
\newblock In \emph{Proceedings of the IEEE/CVF Conference on Computer Vision
  and Pattern Recognition}, pp.\  23751--23762, June 2026{\natexlab{a}}.
\newblock URL \url{https://arxiv.org/abs/2605.24456}.

\bibitem[Li et~al.(2026{\natexlab{b}})Li, Chen, Song, Lei, Zhang, Yan, Pan, and
  Liu]{li2026ipibench}
Jinzhao Li, Yinuo Chen, Wenxuan Song, Yijia Lei, Yichi Zhang, Honglei Yan,
  Panwang Pan, and Miao Liu.
\newblock {IPIBench}: Evaluating interactive proactive intelligence of {MLLMs}
  under continuous streams.
\newblock \emph{arXiv preprint arXiv:2605.27074}, 2026{\natexlab{b}}.
\newblock URL \url{https://arxiv.org/abs/2605.27074}.

\bibitem[Li et~al.(2022)Li, Li, Xiong, and Hoi]{Li2022BLIPBL}
Junnan Li, Dongxu Li, Caiming Xiong, and Steven C.~H. Hoi.
\newblock Blip: Bootstrapping language-image pre-training for unified
  vision-language understanding and generation.
\newblock In \emph{Proceedings of the 39th International Conference on Machine
  Learning}, volume 162 of \emph{Proceedings of Machine Learning Research},
  pp.\  12888--12900. PMLR, 2022.

\bibitem[Li et~al.(2026{\natexlab{c}})Li, Zhang, Long, Chen, Song, Bai, Yang,
  Xie, Yang, Liu, Zhou, and Lin]{li2026qwen3vlembedding}
Mingxin Li, Yanzhao Zhang, Dingkun Long, Keqin Chen, Sibo Song, Shuai Bai,
  Zhibo Yang, Pengjun Xie, An~Yang, Dayiheng Liu, Jingren Zhou, and Junyang
  Lin.
\newblock Qwen3-vl-embedding and qwen3-vl-reranker: A unified framework for
  state-of-the-art multimodal retrieval and ranking.
\newblock arXiv preprint arXiv:2601.04720, 2026{\natexlab{c}}.

\bibitem[Lin et~al.(2025)Lin, Lee, Shoeybi, Lin, Catanzaro, and
  Ping]{Lin2024MMEmbedUM}
Sheng-Chieh Lin, Chankyu Lee, Mohammad Shoeybi, Jimmy Lin, Bryan Catanzaro, and
  Wei Ping.
\newblock {MM-Embed}: Universal multimodal retrieval with multimodal {LLM}s.
\newblock In \emph{The Thirteenth International Conference on Learning
  Representations}, 2025.

\bibitem[Lin et~al.(2014)Lin, Maire, Belongie, Hays, Perona, Ramanan,
  Doll{\'a}r, and Zitnick]{lin2014microsoft}
Tsung-Yi Lin, Michael Maire, Serge Belongie, James Hays, Pietro Perona, Deva
  Ramanan, Piotr Doll{\'a}r, and C.~Lawrence Zitnick.
\newblock Microsoft {COCO}: Common objects in context.
\newblock In \emph{European Conference on Computer Vision}, pp.\  740--755.
  Springer, 2014.

\bibitem[Liu et~al.(2025{\natexlab{a}})Liu, Yang, Gao, Zhu, Zhu, Zhao, and
  Wang]{Liu2025ReasoningGE}
Chunxu Liu, Jiyuan Yang, Ruopeng Gao, Yuhan Zhu, Feng Zhu, Rui Zhao, and Limin
  Wang.
\newblock Reasoning guided embeddings: Leveraging mllm reasoning for improved
  multimodal retrieval.
\newblock \emph{arXiv preprint arXiv:2511.16150}, 2025{\natexlab{a}}.

\bibitem[Liu et~al.(2021{\natexlab{a}})Liu, Wang, Wang, and
  Ordonez]{liu2021visual}
Fuxiao Liu, Yinghan Wang, Tianlu Wang, and Vicente Ordonez.
\newblock Visual news: Benchmark and challenges in news image captioning.
\newblock In \emph{Proceedings of the 2021 Conference on Empirical Methods in
  Natural Language Processing}, pp.\  6761--6771, 2021{\natexlab{a}}.

\bibitem[Liu et~al.(2023)Liu, Feng, Fu, Chen, and Wang]{Liu2023EDISEI}
Siqi Liu, Weixi Feng, Tsu-Jui Fu, Wenhu Chen, and William~Yang Wang.
\newblock Edis: Entity-driven image search over multimodal web content.
\newblock In \emph{Proceedings of the 2023 Conference on Empirical Methods in
  Natural Language Processing}, pp.\  4877--4894, 2023.

\bibitem[Liu et~al.(2025{\natexlab{b}})Liu, Zhang, Cai, Jiang, Hu, Yao, Wang,
  and Xie]{Liu2024LamRALM}
Yikun Liu, Yajie Zhang, Jiayin Cai, Xiaolong Jiang, Yao Hu, Jiangchao Yao,
  Yanfeng Wang, and Weidi Xie.
\newblock Lamra: Large multimodal model as your advanced retrieval assistant.
\newblock In \emph{Proceedings of the IEEE/CVF Conference on Computer Vision
  and Pattern Recognition (CVPR)}, pp.\  4015--4025, June 2025{\natexlab{b}}.

\bibitem[Liu et~al.(2021{\natexlab{b}})Liu, Rodriguez-Opazo, Teney, and
  Gould]{liu2021image}
Zheyuan Liu, Cristian Rodriguez-Opazo, Damien Teney, and Stephen Gould.
\newblock Image retrieval on real-life images with pre-trained
  vision-and-language models.
\newblock In \emph{Proceedings of the IEEE/CVF International Conference on
  Computer Vision}, pp.\  2125--2134, 2021{\natexlab{b}}.

\bibitem[Ma et~al.(2024)Ma, Lin, Li, Chen, and Lin]{Ma2024UnifyingMR}
Xueguang Ma, Sheng-Chieh Lin, Minghan Li, Wenhu Chen, and Jimmy Lin.
\newblock Unifying multimodal retrieval via document screenshot embedding.
\newblock In \emph{Proceedings of the 2024 Conference on Empirical Methods in
  Natural Language Processing}, pp.\  6492--6505, 2024.

\bibitem[Meng et~al.(2026)Meng, Jiang, Liu, Su, Yang, Fu, Qin, Thirukovalluru,
  Zhang, Chen, Xu, Xiong, Zhou, Chen, and Yavuz]{Meng2025VLM2VecV2AM}
Rui Meng, Ziyan Jiang, Ye~Liu, Mingyi Su, Xinyi Yang, Yuepeng Fu, Can Qin,
  Raghuveer Thirukovalluru, Xuan Zhang, Zeyuan Chen, Ran Xu, Caiming Xiong,
  Yingbo Zhou, Wenhu Chen, and Semih Yavuz.
\newblock Vlm2vec-v2: Advancing multimodal embedding for videos, images, and
  visual documents.
\newblock \emph{Transactions on Machine Learning Research}, 2026.

\bibitem[Radford et~al.(2021)Radford, Kim, Hallacy, Ramesh, Goh, Agarwal,
  Sastry, Askell, Mishkin, Clark, Krueger, and
  Sutskever]{Radford2021LearningTV}
Alec Radford, Jong~Wook Kim, Chris Hallacy, Aditya Ramesh, Gabriel Goh,
  Sandhini Agarwal, Girish Sastry, Amanda Askell, Pamela Mishkin, Jack Clark,
  Gretchen Krueger, and Ilya Sutskever.
\newblock Learning transferable visual models from natural language
  supervision.
\newblock In \emph{Proceedings of the 38th International Conference on Machine
  Learning}, volume 139 of \emph{Proceedings of Machine Learning Research},
  pp.\  8748--8763. PMLR, 2021.

\bibitem[Shen et~al.(2025)Shen, Zhao, Gu, Gao, Liu, Huang, Gao, Lin, Zhang, and
  Chen]{NEURIPS2025_b5dc49f4}
Junhao Shen, Haiteng Zhao, Yuzhe Gu, Songyang Gao, Kuikun Liu, Haian Huang,
  Jianfei Gao, Dahua Lin, Wenwei Zhang, and Kai Chen.
\newblock Semi-off-policy reinforcement learning for vision-language
  slow-thinking reasoning.
\newblock In \emph{Advances in Neural Information Processing Systems},
  volume~38, pp.\  125082--125115. Curran Associates, Inc., 2025.
\newblock \doi{10.52202/085713-4169}.
\newblock URL \url{https://arxiv.org/abs/2507.16814}.

\bibitem[van~der Maaten \& Hinton(2008)van~der Maaten and
  Hinton]{Maaten2008TSNE}
Laurens van~der Maaten and Geoffrey~E. Hinton.
\newblock Visualizing data using t-sne.
\newblock \emph{Journal of Machine Learning Research}, 9:\penalty0 2579--2605,
  2008.

\bibitem[Wang et~al.(2023)Wang, Li, Dai, Chen, Zhou, Meng, Zhou, and
  Sun]{Wang2023LabelWA}
Lean Wang, Lei Li, Damai Dai, Deli Chen, Hao Zhou, Fandong Meng, Jie Zhou, and
  Xu~Sun.
\newblock Label words are anchors: An information flow perspective for
  understanding in-context learning.
\newblock In \emph{Proceedings of the 2023 Conference on Empirical Methods in
  Natural Language Processing}, pp.\  9840--9855, 2023.

\bibitem[Wang et~al.(2024)Wang, Bai, Tan, Wang, Fan, Bai, Chen, Liu, Wang, Ge,
  Fan, Dang, Du, Ren, Men, Liu, Zhou, Zhou, and Lin]{Wang2024Qwen2VLEV}
Peng Wang, Shuai Bai, Sinan Tan, Shijie Wang, Zhihao Fan, Jinze Bai, Keqin
  Chen, Xuejing Liu, Jialin Wang, Wenbin Ge, Yang Fan, Kai Dang, Mengfei Du,
  Xuancheng Ren, Rui Men, Dayiheng Liu, Chang Zhou, Jingren Zhou, and Junyang
  Lin.
\newblock Qwen2-vl: Enhancing vision-language model's perception of the world
  at any resolution.
\newblock \emph{arXiv preprint arXiv:2409.12191}, 2024.

\bibitem[Wang \& Isola(2020)Wang and Isola]{Wang2020AlignmentUniformity}
Tongzhou Wang and Phillip Isola.
\newblock Understanding contrastive representation learning through alignment
  and uniformity on the hypersphere.
\newblock In \emph{Proceedings of the 37th International Conference on Machine
  Learning}, volume 119 of \emph{Proceedings of Machine Learning Research},
  pp.\  9929--9939. PMLR, 2020.

\bibitem[Wei et~al.(2024)Wei, Chen, Chen, Hu, Zhang, Fu, Ritter, and
  Chen]{Wei2024UniIRTA}
Cong Wei, Yang Chen, Haonan Chen, Hexiang Hu, Ge~Zhang, Jie Fu, Alan Ritter,
  and Wenhu Chen.
\newblock Uniir: Training and benchmarking universal multimodal information
  retrievers.
\newblock In \emph{Computer Vision -- ECCV 2024}, volume 15145 of \emph{Lecture
  Notes in Computer Science}, pp.\  387--404. Springer, 2024.
\newblock \doi{10.1007/978-3-031-73021-4_23}.

\bibitem[Wu et~al.(2021)Wu, Gao, Guo, Al-Halah, Rennie, Grauman, and
  Feris]{wu2021fashion}
Hui Wu, Yupeng Gao, Xiaoxiao Guo, Ziad Al-Halah, Steven Rennie, Kristen
  Grauman, and Rogerio Feris.
\newblock Fashion iq: A new dataset towards retrieving images by natural
  language feedback.
\newblock In \emph{Proceedings of the IEEE/CVF Conference on Computer Vision
  and Pattern Recognition}, pp.\  11307--11317, 2021.

\bibitem[Wu et~al.(2026{\natexlab{a}})Wu, Mei, Ma, Chai, Lan, Zhao, Yan, Chen,
  Hu, Peng, Lin, Zhou, Yin, Wang, Rao, Lv, Li, and Sun]{Wu2026RIME}
Peixi Wu, Ke~Mei, Feipeng Ma, Bosong Chai, Zhibin Lan, Chenxi Zhao, Shannan
  Yan, Jie Chen, Zhangchi Hu, Yansong Peng, Bo~Lin, Junjie Zhou, Dacheng Yin,
  Tianyi Wang, Fengyun Rao, Jing Lv, Hebei Li, and Xiaoyan Sun.
\newblock Beyond chain-of-thought: Rewrite as a universal interface for
  generative multimodal embeddings.
\newblock \emph{arXiv preprint arXiv:2604.22280}, 2026{\natexlab{a}}.

\bibitem[Wu et~al.(2026{\natexlab{b}})Wu, Yang, Ma, Chai, Lin, Yuan, Yang, Gao,
  Li, and Sun]{Wu2026LaME}
Peixi Wu, Biao Yang, Feipeng Ma, Bosong Chai, Bo~Lin, Wei Yuan, Fan Yang,
  Tingting Gao, Hebei Li, and Xiaoyan Sun.
\newblock Lame: Learning to think in latent space for multimodal embedding via
  information bottleneck.
\newblock \emph{arXiv preprint arXiv:2606.13061}, 2026{\natexlab{b}}.

\bibitem[Xiao et~al.(2024)Xiao, Tian, Chen, Han, and
  Lewis]{Xiao2023EfficientSL}
Guangxuan Xiao, Yuandong Tian, Beidi Chen, Song Han, and Mike Lewis.
\newblock Efficient streaming language models with attention sinks.
\newblock In \emph{International Conference on Learning Representations}, 2024.

\bibitem[Xiao et~al.(2026)Xiao, Ma, Gu, cheng Jason~Chen, Chen,
  Ord{\'o}{\~n}ez, and Mohan]{Xiao2025MetaEmbedSM}
Zilin Xiao, Qi~Ma, Mengting Gu, Chun cheng Jason~Chen, Xintao Chen, Vicente
  Ord{\'o}{\~n}ez, and Vijai Mohan.
\newblock Metaembed: Scaling multimodal retrieval at test-time with flexible
  late interaction.
\newblock In \emph{The Fourteenth International Conference on Learning
  Representations}, 2026.

\bibitem[Xie et~al.(2026)Xie, Zhao, Liu, Zhu, Chen, Ye, Chen, Xu, Dong, Wang,
  Xu, Shi, Wu, Zhang, Shao, Chang, Duan, and Wang]{xie2026step}
Weichu Xie, Haozhe Zhao, Wenpu Liu, Yongfu Zhu, Liang Chen, Minghao Ye, Zirong
  Chen, Yuqi Xu, Shuai Dong, Ziyue Wang, Xinbo Xu, Kean Shi, Ruoyu Wu, Xiaoying
  Zhang, Wenqi Shao, Baobao Chang, Nan Duan, and Jiaqi Wang.
\newblock Step-wise rubric rewards for {LLM} reasoning.
\newblock \emph{arXiv preprint arXiv:2605.17291}, 2026.
\newblock URL \url{https://arxiv.org/abs/2605.17291}.

\bibitem[Yao et~al.(2026)Yao, Xu, Guo, Han, Yang, Zhang, Zhang, Zeng, and
  Liu]{yao2026acerouter}
Zhiyuan Yao, Zishan Xu, Yifu Guo, Zhiguang Han, Cheng Yang, Shuo Zhang, Weinan
  Zhang, Xingshan Zeng, and Weiwen Liu.
\newblock {ACE-Router}: Generalizing history-aware routing from {MCP} tools to
  the agent web.
\newblock In \emph{Proceedings of the 64th Annual Meeting of the Association
  for Computational Linguistics (Volume 1: Long Papers)}, pp.\  6224--6240.
  Association for Computational Linguistics, July 2026.
\newblock \doi{10.18653/v1/2026.acl-long.281}.
\newblock URL \url{https://aclanthology.org/2026.acl-long.281/}.

\bibitem[Zeng et~al.(2026)Zeng, Xu, Zhang, Wu, Wang, Lin, Li, Wang, Shang,
  Jiang, Zhang, Yu, Liu, and Liu]{zeng2026agenticdata}
Xingshan Zeng, Zishan Xu, Boju Zhang, Yuzhou Wu, Lingzhi Wang, Jianghao Lin,
  Liangyou Li, Yasheng Wang, Lifeng Shang, Xin Jiang, Weinan Zhang, Yong Yu,
  Qun Liu, and Weiwen Liu.
\newblock What makes good agentic data? an {ACE} lens on data generation for
  {LLM} agents.
\newblock \emph{arXiv preprint arXiv:2608.27260}, 2026.
\newblock URL \url{https://arxiv.org/abs/2608.27260}.

\bibitem[Zhai et~al.(2023)Zhai, Mustafa, Kolesnikov, and
  Beyer]{Zhai2023SigmoidLF}
Xiaohua Zhai, Basil Mustafa, Alexander Kolesnikov, and Lucas Beyer.
\newblock Sigmoid loss for language image pre-training.
\newblock In \emph{Proceedings of the IEEE/CVF International Conference on
  Computer Vision (ICCV)}, pp.\  11975--11986, October 2023.

\bibitem[Zhang et~al.(2026{\natexlab{a}})Zhang, Liu, Zhao, Hou, Zhang, Xie,
  Liu, and Li]{gameverse2026}
Kuan Zhang, Dongchen Liu, Qiyue Zhao, Jinkun Hou, Xinran Zhang, Qinlei Xie,
  Miao Liu, and Yiming Li.
\newblock {GameVerse}: Can vision-language models learn from video-based
  reflection?
\newblock \emph{arXiv preprint arXiv:2603.06656}, 2026{\natexlab{a}}.
\newblock URL \url{https://arxiv.org/abs/2603.06656}.

\bibitem[Zhang et~al.(2026{\natexlab{b}})Zhang, Liu, Zhao, Xin, Su, Wang, Yin,
  Ma, Li, Gu, Wu, Zhang, Li, Chen, and Li]{zhang2026towards}
Kuan Zhang, Dongchen Liu, Qiyue Zhao, Tianyu Xin, Yue Su, Haisheng Wang, Han
  Yin, Hongbo Ma, Peize Li, Tianjun Gu, Xiangnan Wu, Xinran Zhang, Yongxuan Li,
  Zirong Chen, and Yiming Li.
\newblock Towards generalist game players: An investigation of foundation
  models in the game multiverse.
\newblock \emph{arXiv preprint arXiv:2605.09965}, 2026{\natexlab{b}}.
\newblock URL \url{https://arxiv.org/abs/2605.09965}.

\bibitem[Zhang et~al.(2025)Zhang, Zhang, Xie, Li, Dai, Long, Xie, Zhang, Li,
  and Zhang]{Zhang2024GMEIU}
Xin Zhang, Yanzhao Zhang, Wen Xie, Mingxin Li, Ziqi Dai, Dingkun Long, Pengjun
  Xie, Meishan Zhang, Wenjie Li, and Min Zhang.
\newblock Bridging modalities: Improving universal multimodal retrieval by
  multimodal large language models.
\newblock In \emph{Proceedings of the IEEE/CVF Conference on Computer Vision
  and Pattern Recognition (CVPR)}, pp.\  9274--9285, June 2025.

\end{thebibliography}
\end{document}